\documentclass[12pt]{article}

\usepackage{arxiv}

\usepackage[utf8]{inputenc} 
\usepackage[T1]{fontenc}    
\usepackage{hyperref}       
\usepackage{url}            
\usepackage{booktabs}       
\usepackage{amsfonts}       
\usepackage{amsmath}
\usepackage{nicefrac}       
\usepackage{microtype}      
\usepackage{cleveref}       
\usepackage{lipsum}         
\usepackage{graphicx}
\usepackage{natbib}
\usepackage{doi}

\usepackage{graphicx}
\usepackage{booktabs}
\usepackage{multirow}
\usepackage{setspace}
\usepackage{makecell}
\usepackage{amssymb}
\usepackage{tabularx}
\usepackage{caption}
\usepackage{subcaption}
\usepackage[table]{xcolor}
\usepackage{wrapfig}
\usepackage{fontawesome5}

\title{FineViT: Progressively Unlocking Fine-Grained Perception with Dense Recaptions}

\date{}

\author{Peisen Zhao \hspace{0.4cm} Xiaopeng Zhang\textsuperscript{*} \hspace{0.4cm} Mingxing Xu \hspace{0.4cm} Ruoyu Sun \hspace{0.4cm} Zewei Du \hspace{0.4cm} Dunzheng Wang \\ \\ Guanghao Zheng \hspace{0.4cm} Haohang Xu \hspace{0.4cm} Zhibo Zhang \hspace{0.4cm} Yuhang Zhang \hspace{0.4cm} Yi Ai \hspace{0.4cm} Lin Liu \hspace{0.4cm} Qi Tian\textsuperscript{\faEnvelope[regular]} \\ \\
{\makebox[\textwidth][c]{\textit{Huawei Inc.}}}
}

\renewcommand{\headeright}{Technical Report}
\renewcommand{\undertitle}{~}
\renewcommand{\shorttitle}{\textit{FineViT}}

\begin{document}
\maketitle

\insert\footins{\noindent\footnotesize \thanks{* Project Lead,} \, \faEnvelope[regular] Corresponding Author: \texttt{tian.qi1@huawei.com}, \faGithub Repo: https://github.com/PeisenZhao/FineViT}

\begin{abstract}
While Multimodal Large Language Models (MLLMs) have experienced rapid advancements, their visual encoders frequently remain a performance bottleneck. Conventional CLIP-based encoders struggle with dense spatial tasks due to the loss of visual details caused by low-resolution pretraining and the reliance on noisy, coarse web-crawled image-text pairs. To overcome these limitations, we introduce \textbf{FineViT}, a novel vision encoder specifically designed to unlock fine-grained perception. By replacing coarse web data with dense recaptions, we systematically mitigate information loss through a progressive training paradigm.: first, the encoder is trained from scratch at a high native resolution on billions of global recaptioned image-text pairs, establishing a robust, detail rich semantic foundation. Subsequently, we further enhance its local perception through LLM alignment, utilizing our curated \textbf{FineCap-450M} dataset that comprises over $450$ million high quality local captions. Extensive experiments validate the effectiveness of the progressive strategy. FineViT achieves state-of-the-art zero-shot recognition and retrieval performance, especially in long-context retrieval, and consistently outperforms multimodal visual encoders such as SigLIP2 and Qwen-ViT when integrated into MLLMs. We hope FineViT could serve as a powerful new baseline for fine-grained visual perception.
  \keywords{vision encoder \and fine-grained perception \and dense recaptions}
\end{abstract}

\section{Introduction}
\label{sec:intro}
The rapid evolution of Multimodal Large Language Models (MLLMs) has fundamentally altered the landscape of artificial intelligence. As the reasoning capabilities of Large Language Models (LLMs) have scaled exponentially, the visual component of MLLMs has frequently lagged behind, serving as a bottleneck to overall performance. For a significant period, the visual encoder of MLLMs was treated as a static artifact, typically a pre-trained Vision Transformer (ViT) such as CLIP \cite{radford2021learning}, and connected to the LLMs via a lightweight adapter \cite{liu2024llavanext}. While this paradigm facilitates rapid convergence, it imposes severe structural limitations due to the post alignment of different pretrained modalities, and suffer limited performance especially for spatial visual tasks.

The core challenge for designing visual encoders in MLLMs lies in the \textbf{loss of detailed information} inherent in current CLIP-based models and the \textbf{modality gap} dilemma when aligning visual models to well-trained LLMs. As for the loss of detailed information, traditional visual encoders, designed primarily for image classification or retrieval, enforce rigid constraints on input resolution, typically resizing images to fixed squares (\emph{e.g.}, $224 \times 224$ or $336 \times 336$). This preprocessing step, while computationally convenient for batch processing, functions as a lossy compression algorithm that obliterates the fine details necessary for dense tasks such as Optical Character Recognition (OCR) and small object grounding. Furthermore, information loss is exacerbated by the reliance on low-quality annotated data. Since CLIP-based models require large-scale image-text pairs, where the data is usually sourced through web crawling with simple filtering mechanisms. Such an approach has facilitated large-scale data collection, exemplified by collections like LAION-400M~\cite{schuhmann2021laion}, it has inadvertently compromised data quality. These internet crawled pairs frequently exhibit misalignments between images and their corresponding textual content, and the textual descriptions are often brief and lack detailed information. Finally, the modality alignment challenge is exacerbated by a fundamental mismatch in training objectives, where visual encoders are typically pre-trained using contrastive learning, which focuses on global semantic alignment, whereas LLMs are optimized via an autoregressive objective. This discrepancy creates a fragmented training paradigm where the pre-training and alignment phases operate under different principles. Consequently, it makes the alignment process opaque and particularly challenging when scaling to large LLMs.
\begin{figure}[t]
    \centering
    \begin{subfigure}[b]{0.66\textwidth}
        \centering
        \includegraphics[width=\textwidth]{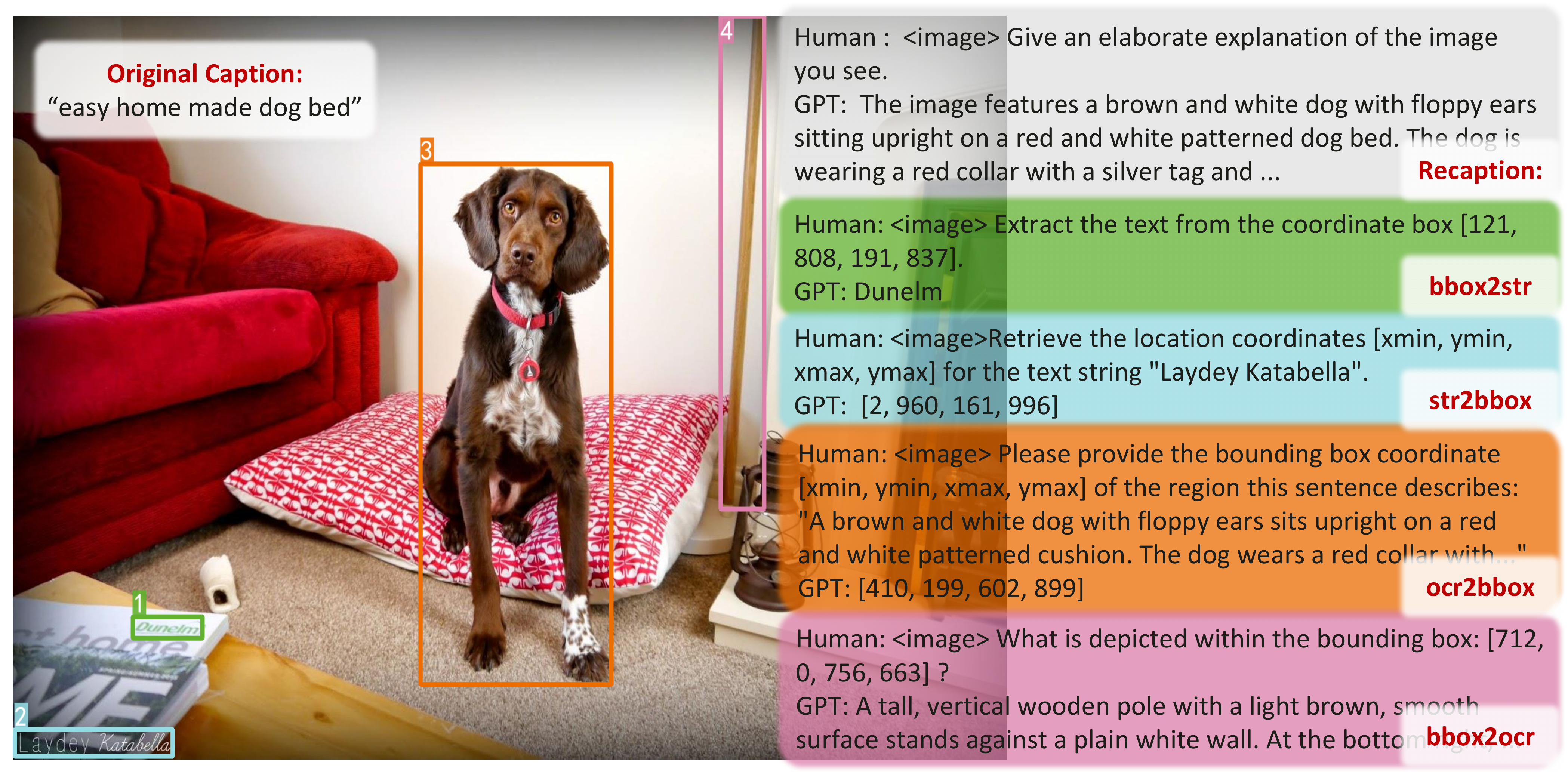}
        \caption{}
        \label{fig:label_a}
    \end{subfigure}
    \hfill 
    \begin{subfigure}[b]{0.33\textwidth}
        \centering
        \includegraphics[width=\textwidth]{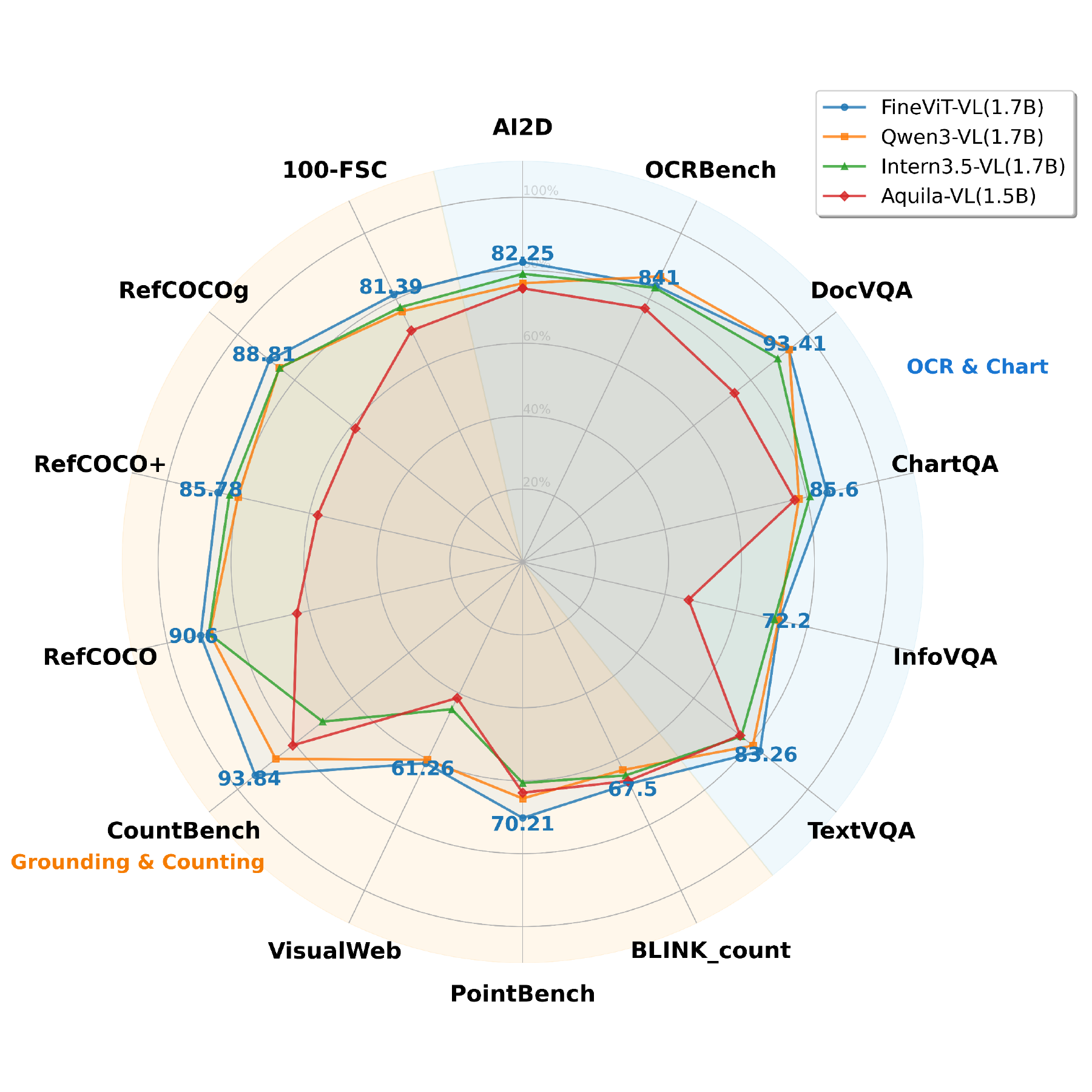}
        \caption{}
        \label{fig:label_b}
    \end{subfigure}
    \caption{(a) Construction of rich QA training data from FineCap-450M, highlighting detailed local-level tasks. (b) FineViT performance boost in multimodal tasks enabled by FineCap-450M local-task pre-training.}
    \label{fig:motivation}
\end{figure}

To address these challenges, this paper revisits the design principles of visual foundation models and introduces \textbf{FineViT}, a top leading vision encoder \textbf{trained from scratch} and specifically tailored for \textbf{fine-grained perception}. Our primary contribution is a progressive training recipe paired with large-scale, fine-grained recaptioned data, which endows the encoder with robust feature representation while embedding precise perception. Specifically, the encoder is first trained with large scale contrastive learning at a $448 \times 448$ native-resolution on \textbf{billions of recaptioned image-text pairs}. Unlike raw captioned datasets such as LAION-400M \cite{schuhmann2021laion}, our recaptioned data provide dense descriptive detail, while the higher resolution preserves critical spatial information during encoding. Consequently, FineViT achieves exceptional zero-shot recognition and retrieval performance relative to CLIP-based counterparts. To further empower the fine-grained capabilities required by MLLMs, we align the encoder with LLMs using our curated \textbf{FineCap-450M dataset, comprising over 450 million local captions} spanning natural scenes and diverse text-rich environments. To the best of our knowledge, this represents the largest fine-grained annotated dataset to date. As illustrated in Figure~\ref{fig:motivation}, equipped with this high-quality data and a robust training strategy, FineViT matches the zero-shot classification and retrieval performance of current SOTA encoders like SigLIP2 \cite{tschannen2025siglip} and Seed-ViT \cite{guo2025seed1}. When integrated into MLLMs, FineViT outperforms multimodal models such as Qwen3-VL~\cite{Qwen3-VL} and Intern3.5-VL~\cite{wang2025internvl3}. We release FineViT to the community as a powerful baseline for fine-grained visual perception.

The core contributions of this paper are summarized below:

$\bullet$ We introduce FineViT, a trained from scratch visual encoder at a native, high resolutional training paradigm, and progressively unlocking the fine-grained perception capabilities.

$\bullet$ To drive the training of FineViT, we construct FineCap-450M, the largest fine-grained dataset to date which comprise over 450 million high quality region captions.

$\bullet$ FineViT matches SOTA models like SigLIP2~\cite{tschannen2025siglip} in zero-shot tasks and excels in long-context retrieval. When integrated into MLLMs, it consistently outperforms multimodal encoders like Qwen-ViT and InternViT.




\section{Related Work}
\subsection{Visual Foundation Models}
Early MLLMs predominantly rely on off-the-shelf visual encoders, such as CLIP \cite{radford2021learning}, to align visual and textual semantic spaces. The widely adopted CLIP visual backbone \cite{radford2021learning}, which aligns image and text representations by maximizing the cosine similarity of matched pairs in a shared embedding space, has proven more effective than previous self-supervised models \cite{he2020momentum} that lack semantic alignment with text. To address the inefficiency of training large-scale CLIP models from scratch, EVA-CLIP \cite{sun2023eva, sun2024eva} bridges the gap between generative self-supervised learning and discriminative semantic alignment. It achieves this by regressing the features of a pre-trained CLIP model from masked image patches, allowing for efficient scaling to billions of parameters. Furthermore, recognizing that scaling CLIP is bottlenecked by the pairwise softmax loss, which requires large batch sizes for stable training, SigLIP \cite{zhai2023sigmoid} replaces this objective with a simple pairwise sigmoid loss. This decouples the batch size from the objective, enabling efficient scaling and improved zero-shot performance. Most recently, SigLIP2 \cite{tschannen2025siglip} addresses the limited spatial precision of these contrastive models by incorporating auxiliary dense supervision losses, including masked prediction and self-distillation, forcing the encoder to retain local features.
\subsection{Vision Models Evolution in MLLMs}
To overcome the modality gap and resolution constraints of the CLIP series, recent state-of-the-art MLLMs increasingly employ visual encoders trained from scratch. These models prioritize native resolution processing, dense supervision, and deep LLM alignment. For instance, InternViT \cite{chen2024internvl, chen2024expanding, zhu2025internvl3} adopts a continuous learning paradigm, progressively aligning the encoder as a trainable prefix to handle high resolution inputs (e.g., 448px tiles) and dense OCR tasks. Similarly, the Qwen-ViT series \cite{bai2025qwen2} utilizes a Naive Resolution mechanism to process images at their native aspect ratios. By converting images into variable-length token sequences and incorporating window attention with 2D-RoPE, it efficiently mitigates the quadratic computational costs of high-resolution attention. Meanwhile, Seed-ViT \cite{guo2025seed1} emphasizes data quality, pre-training on massive interleaved corpora and detailed synthetic recaptions rather than noisy web text. Challenging standard contrastive approaches, AIMv2 \cite{fini2025multimodal} introduces a visual encoder pre-trained via a multimodal autoregressive objective. By simultaneously predicting the next image patch and text token, AIMv2 ensures token level dense supervision, yielding highly robust features for localization and recognition.
\label{sec:related}

\section{Methods}


\subsection{Model Architecture} 
Following the architecture of native-resolution visual encoders~\cite{guo2025seed1,bai2025qwen2}, we employ 2D Rotary Positional Embeddings~\cite{su2024roformer} (2D RoPE) to encode the positional information of the input patches, setting the patch size to $14 \times 14$. To facilitate efficient spatial compression in subsequent stages, input images are resized to the nearest resolution with a multiple of $28 \times 28$ pixels. By stacking multiple Transformer blocks composed of attention mechanisms and Feed-Forward Networks (FFN), we construct a $28$-layer visual encoder with approximately $0.86$ billion parameters. The detailed architecture is presented in Table~\ref{tab:model_architecture}.

\subsection{Training Recipe}
\label{subsec:training_rec}
\begin{table*}[t]
    \centering
    \caption{Architectural details of FineViT.}
    \label{tab:model_architecture}
    \resizebox{1.0\textwidth}{!}{
    \setlength{\tabcolsep}{6pt} 
    \begin{tabular}{lccccccc}
        \toprule
        \textbf{Model} & \textbf{Depth} & \textbf{Patch Size} & \textbf{Hidden Size} & \textbf{Interm. Size} & \textbf{Heads} & \textbf{Activation} & \textbf{Pos. Embed} \\
        \midrule
        FineViT & 28 & 14 & 1536 & 4608 & 16 & SiLU & 2D RoPE \\
        \bottomrule
    \end{tabular}
    }
\end{table*}

\begin{figure}[t]
  \centering
  \includegraphics[width=1.0\columnwidth]{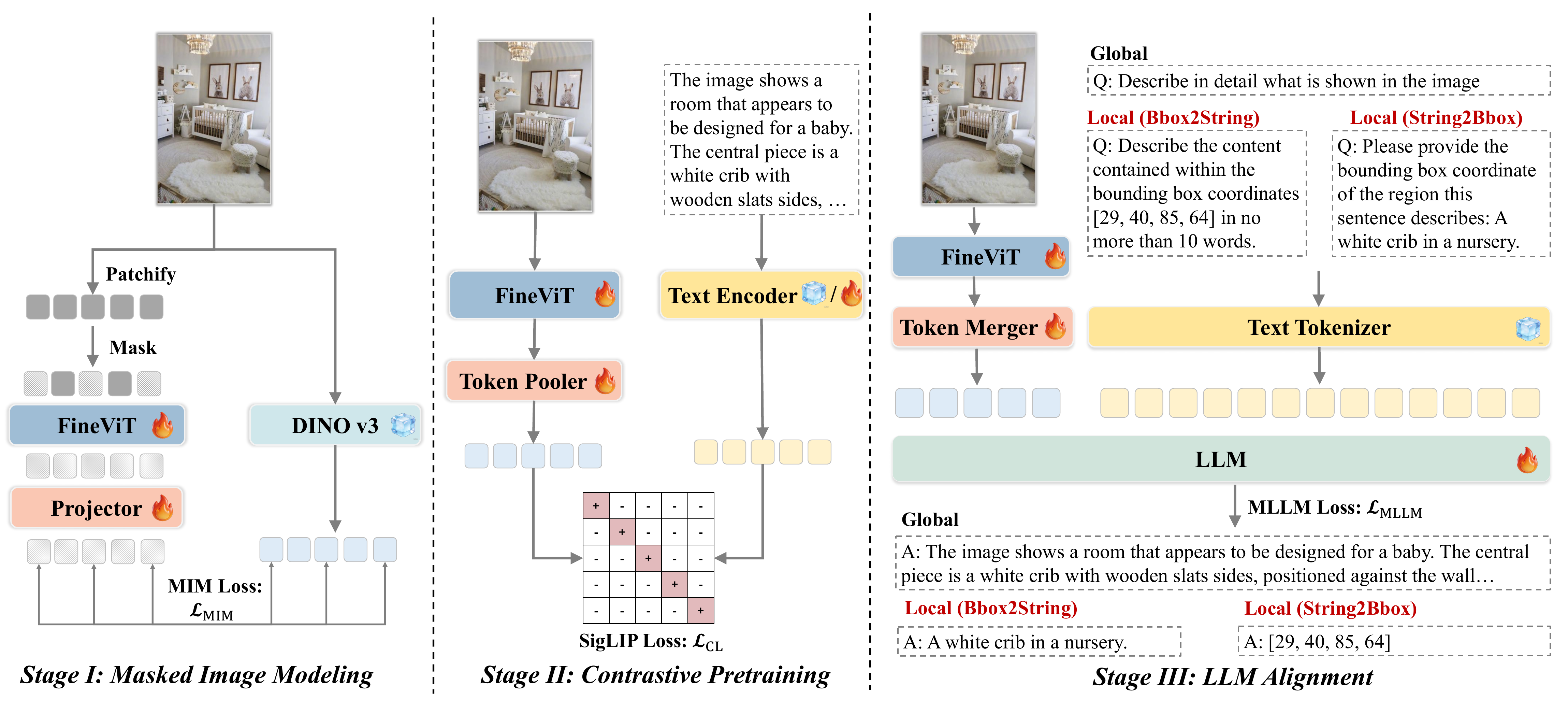}
  \caption{Illustration of FineViT training stages. Stage I utilizes a Masked Image Modeling (MIM) loss to establish foundational visual perception. Stage II implements a SigLIP loss via large-scale image-text contrastive learning to bridge the gap between visual features and semantic concepts. Finally, Stage III integrates a Large Language Model (LLM) at high resolution to train the model on global and local Question-Answering (QA) tasks, ensuring a robust sensitivity to fine-grained visual details. }
  \label{fig:training}
\end{figure}

To mitigate biases introduced by pretrained models, FineViT is trained from scratch. Contemporary training paradigms for visual encoders typically fall into two categories: self-supervised learning via visual pretext tasks (\textit{e.g.}, masked image modeling~\cite{he2022masked}, next scale prediction~\cite{tian2024visual}) and language supervised learning (\textit{e.g.}, image text contrastive learning~\cite{radford2021learning}, autoregressive pretraining~\cite{fini2025multimodal}). To stabilize the training process and systematically investigate encoder properties across different training stages, we employ a three stage curriculum learning strategy, \textit{i.e.}, MIM initialization, large scale contrastive learning, and LLM alignment. This strategy progresses from establishing foundational visual perception via pretext tasks to achieving complex visual understanding through LLM alignment. Furthermore, we progressively scale the input resolution throughout these stages to enhance the model's ability to capture fine-grained visual details. The training stages are illustrated in Figure~\ref{fig:training} and detailed below.

\noindent\textbf{Stage I: Initialization with MIM.} 
In the initial phase, we employ Masked Image Modeling (MIM) to pretrain the visual encoder. This process equips the model with preliminary spatial awareness and geometric reasoning capabilities through the reconstruction of masked regions. Following EVA~\cite{fang2023eva} and MVP~\cite{wei2022mvp}, we define the reconstruction target within the feature space to enhance learning efficiency. To isolate the visual encoder's learning process from label and language supervision, we conduct this phase entirely on a large-scale, unlabeled image dataset. Specifically, we utilize DINOv3~\cite{simeoni2025dinov3}, a robust self-supervised vision model, as the teacher network $\mathcal{T}$ to provide the target features. Following a $75\%$ masking strategy at $256 \times 256$ resolution, the model is optimized via the loss function in Eq.~\ref{eq:stage1}. Here, $\mathcal{M}$ refers to the masked patches, and $\Phi(\cdot)$ represents our visual encoder:

\begin{equation}
\label{eq:stage1}
\mathcal{L}_{\text{MIM}} = \sum_{i \in \mathcal{M}} \left\| \Phi(x)_i - \mathcal{T}(x)_i \right\|^2_2
\end{equation}

\noindent\textbf{Stage II: Large Scale Contrastive Learning.} 
During this phase, we introduce text supervision and employ large scale contrastive learning to further optimize the visual encoder $\Phi(\cdot)$. Specifically, the visual branch is initialized with the MIM pretrained model from Stage I, while the text branch is initialized using the weights of the SigLIP2~\cite{tschannen2025siglip} Giant model. Given that the text encoder is already well-aligned with visual features, we initially freeze its parameters and selectively unfreeze them for joint fine-tuning in later steps.
In contrast to SigLIP2, which incorporates native resolution only during the final stage of training, \textit{FineViT maintains native resolution throughout the entire contrastive learning process}. This approach allows the model to inherently adapt to image data with diverse resolutions and aspect ratios. As training progresses, the maximum resolution is scaled from $336^{2}$ to $448^{2}$ pixels.
Furthermore, to overcome the context limitations of standard CLIP-based models (\textit{e.g.}, $64$–$77$ tokens), which hinder the processing of detailed recaptioned datasets, we progressively extend the text context length from $64$ to $256$ tokens. The Stage II objective is defined in Eq.~\ref{eq:stage2}, where $z_{i,j}$ denotes the pairwise similarity between image $x_i$ and text $t_j$, and $y_{i,j} \in \{1, -1\}$ indicates whether the pair is a positive or negative match.
\begin{equation}
\mathcal{L}_{\text{CL}} = -\frac{1}{n} \sum_{i=1}^n \sum_{j=1}^n \log \left( \sigma(z_{i,j} \cdot y_{i,j}) \right), \quad z_{i,j} = \tau ( \Phi(x_i) \cdot \Psi(t_j) + b )
\label{eq:stage2}
\end{equation}

\noindent\textbf{Stage III: LLM Autoregrresion with Multi-Granularity Alignment.}
In Stage III, we employ a generative training objective to integrate visual features into the MLLM architecture. Beyond global image-text alignment, we enhance fine-grained perception by leveraging over $450$ million region-level QA samples as shown in Figure~\ref{fig:motivation}(a). Specifically, tasks such as \textit{bbox-to-string} and \textit{string-to-bbox} are reformulated into a unified sequence-to-sequence format, enabling the model to interconvert between region captions and spatial coordinates. Besides \textit{bbox-to-ocr} and \textit{ocr-to-bbox} are also designed for training character recognition capabilities.
To capture intricate details, the input resolution is scaled to $1$K. As formulated in Eq.~\ref{eq:stage3}, the model is optimized via a standard autoregressive loss, where the visual features from $\Phi(x)$ are mapped into the language space through a trainable projector $g(\cdot)$. $y_i$ represents the target tokens.
\begin{equation}
\mathcal{L}_{\text{MLLM}} = - \sum_{i=1}^{L} \log P\left(y_i \mid y_{<i}, g\left(\Phi(x)\right)\right)
\label{eq:stage3}
\end{equation}

\subsection{Training Data}
\begin{figure}[t]
  \centering
  \includegraphics[width=1.0\columnwidth]{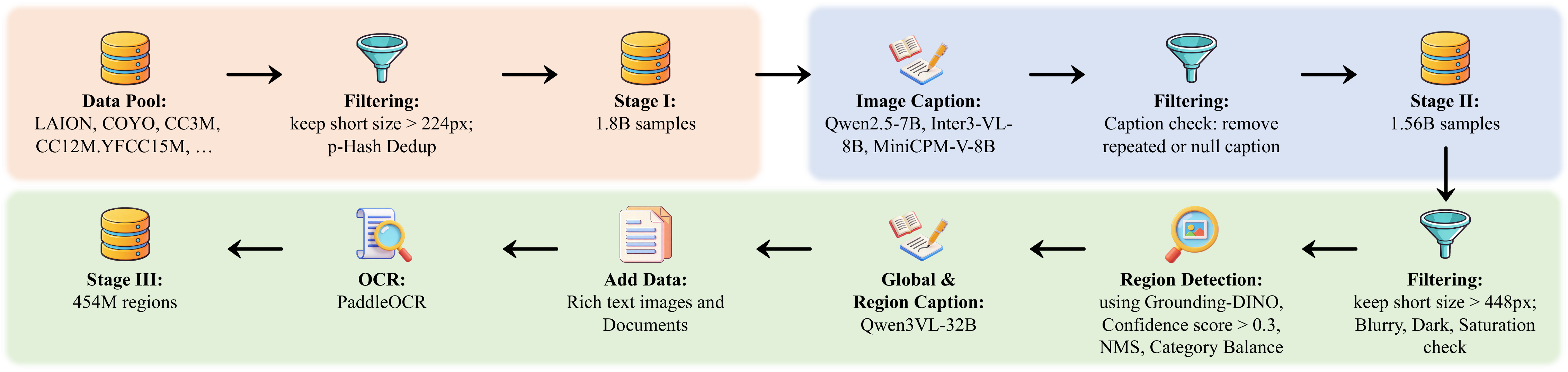}
  \caption{The pipeline of data curation and recaption.}
  \label{fig:training_data}
\end{figure}
In this section, we provide a detailed explaination of the data used at different training stages. As noted in \cite{li2024if}, current large scale web crawled data such as LAION-400M~\cite{schuhmann2021laion} frequently exhibit misaliged image-text pairs, and often, the textual descriptions are brief and lack detailed information, which is limited for fine-grained perception tasks. In our data pipeline, instead of using its raw captions, \textbf{all image-text pairs are recaptioned}, while the caption granularity and quality are progressively improved as the training stages proceed. Detailed descriptions of the data are provided in the appendix.

\noindent\textbf{Data Source.} 
For source data, we collect a diversity of the open-source dataset, covering  MS COCO~\cite{coco}, CC3M~\cite{cc3m}, CC12M~\cite{changpinyo2021cc12m}, YFCC15M~\cite{radford2021learning}, LAION~\cite{schuhmann2021laion}, wukong~\cite{wukong}, datacomp~\cite{gadre2023datacomp}, COYO~\cite{kakaobrain2022coyo-700m}, SBU~\cite{sbu} and Flickr30k~\cite{flickr} \textit{etc.}, as well as extra data that crawled from the internet. After simple filtering and deduplication, where images with shorter size less than $224$ are abandoned, and the nearly-identical samples are deduplicated based on the p-Hash value. \textit{As a result, we conclude with around $1.8$B unlabeled samples, which are used for Stage I MIM initialization training}.

\noindent\textbf{Global Recaption Data.} The raw caption data are usually noisy, and often lack detailed information of the image. To obtain detailed annotations, the collected $1.8B$ samples are feeded into MLLMs for recaption. To ensure diversity and avoid the recaptioner bias a single model, we randomly recaption the billion level data with three models, \emph{i.e.,} Qwen2.5-7B \cite{bai2025qwen2}, Intern3-VL-8B\cite{zhu2025internvl3}, and MiniCPM-V-8B\cite{yao2024minicpm}. The recaptions further undergo a rule based filtering, we check the recaptions and remove those samples with repetitions and null captions. 
This produces $1.56$B recaptioned samples. \textit{The recaptioned $1.56$B samples are used for Stage II large scale contrastive learning.} 

\noindent\textbf{Multi-Granulaity Recaption Data FineCap-450M.}
To empower the model with fine-grained perception ability, we enhance the annotation granularity to region level, and deliberately design a region annotation pipeline for stage III training. The while pipeline is illustrated in Figure~\ref{fig:training_data}

$\bullet$ \textbf{High quality images.} We further sample high-quality images from a pool of $1.56$B candidates. The filtering process initially removes images with a short side smaller than 448 pixels. Following \cite{wu2025qwen}, we discard blurry or out-of-focus images, excessively bright or dark images, and unnaturally high color saturation images, This step roughly excludes $3\%$ unqualified images. Beyond natural images, the dataset also incorporates document-like samples as described in \cite{zheng2025granvit}.

$\bullet$ \textbf{Multi-granulaity recaption.} We further annotate the high quality images at both the global and region levels. At this stage, we utilize Qwen3-VL 32B for recaptioning to optimally balance computational complexity and performance. For region level annotation, we first extract nouns from the global recaptions and use Grounding DINO~\cite{liu2024grounding} to generate candidate bounding boxes. We filter out candidate boxes with confidence scores below $0.3$ and apply non-maximum suppression (NMS) to merge the remaining ones. To maintain category balance, we sample the bounding boxes based on their frequencies. The remaining boxes are then utilized for local captioning. To better incorporate contextual information, we follow \cite{shi2025scaling} and generate local captions conditioned on the global context. Specifically, we input both the local crop and the entire global image into MLLM, prompting it to describe the local details using the global context. Ultimately, this pipeline yields a total of $226$ million annotated regions covering over $600,000$ categories.

$\bullet$ \textbf{Rich text images and Documents.} We notice that for rich text regions in natural images, the local captions often return inaccurate annotations or simply as "text" caption. To enhance the text recognition ability in natural images, we use PaddleOCR \cite{cui2025paddleocr} to detect text, and replace the captions obtained with MLLMs. While for document images, we also utilize PaddleOCR for localized text detection and add it to the dataset.

\begin{table}[t]
    \centering
    \caption{Detailed information for FineCap-450M.}
    \label{tab:training_data}
    \resizebox{1.0\textwidth}{!}{%
    \setlength{\tabcolsep}{3pt} 
    \renewcommand{\arraystretch}{1.4}
    \begin{tabular}{cc ccc cccc | c}
        \toprule
        \multicolumn{2}{c}{Global Caption} & \multicolumn{3}{c}{Local Caption} & \multicolumn{2}{c}{Rich-text OCR } & \multicolumn{2}{c|}{Doc OCR} & \multirow{2}{*}{\makecell{\# total \\ regions}} \\
        \cmidrule(lr){1-2} \cmidrule(lr){3-5} \cmidrule(lr){6-7} \cmidrule(lr){8-9}
        \# images & avg. token & \# regions & avg. token & \# categories & \# regions & avg. token & \# regions & avg. token & \\
        \midrule
        63M & 211.25 & 226M & 56.58 & 631,252 & 142M & 3.88 & 86M & 4.33 & 454M \\
        \bottomrule
    \end{tabular}}
\end{table}

\noindent\textbf{Dataset Details.} As a result, we have annotated global descriptions for $63$ million images, alongside more than $450$ million regions spanning natural, text-rich, and document-only scenarios, noted as \textbf{FineCap-450M}. The detailed data summarization is shown in Table~\ref{tab:training_data}. As far as we know, FineCap-450M is the largest recaptioned data covering fine-grained annotations.

\label{sec:methods}

\section{Experiments}
\label{sec:experiments}

This section begins by detailing the implementation details and evaluation benchmarks for FineViT. Then we systematically assess the performance of the proposed model across each training stage. To demonstrate its comparative advantage, we evaluate FineViT against the SigLIP2 baseline and investigate its cross model adaptability by integrating it with LLMs of varying scales.

\subsection{Experimental Setup}

\begin{table}[t]
    \centering
    \caption{Overview of FineViT training details.}
    \label{tab:training}
    \resizebox{1.0\textwidth}{!}{%
    \setlength{\tabcolsep}{6pt} 
    \begin{tabular}{l c c c}
        \toprule
        \textbf{Training Stage} & \textbf{\makecell{Stage I: MIM}} & \textbf{\makecell{Stage II: Contrastive Learning}} & \textbf{\makecell{Stage III: LLM Align}} \\
        \midrule
        Samples Seen & 1.8B & 9.3B & 0.5B \\
        Batch Size & 4,096 & 49,152 & 4,096 \\
        Resolution & $256 \times 256$ & within $448 \times 448$ & within $1000 \times 1000$ \\
        Learning Rate & $1.0 \times 10^{-3}$ & $1.0 \times 10^{-4}$ & $1.0 \times 10^{-5}$ \\
        Training Modules & ViT & ViT, text encoder & ViT, projector, LLM \\
        \bottomrule
    \end{tabular}}
\end{table}


\noindent\textbf{Implementation Details.} The training pipeline for FineViT follows a progressive three stage paradigm, as detailed in Table \ref{tab:training}. Stage I performs self-supervised pretraining via MIM on $1.8$B samples with a batch size of $4,096$ to establish foundational representations. Stage II transitions to large scale contrastive learning to align visual features with linguistic semantics, processing approximately $9.3$B samples with a batch size of $49,152$. Finally, Stage III targets LLM alignment by scaling the visual input to $1$K resolution, applying a refined learning rate of $1.0 \times 10^{-5}$. All experiments utilize DeepSpeed ZeRO-2 with BF16 precision, employing the AdamW optimizer and a cosine learning rate scheduler. For multimodal understanding evaluation, the FineViT-VL model is trained on a curated dataset of $80$M high quality SFT samples. This dataset is merged from LLaVA-OneVision-1.5 \cite{LLaVA-OneVision-1.5}, FineVision \cite{wiedmann2025finevision}, and Eagle2 \cite{li2025eagle2buildingposttraining}, with a $6$M subset utilized for ablation studies. \textit{Comprehensive details regarding the data composition and training protocols are provided in the appendix. }

\noindent\textbf{Evaluation Benchmarks.} We conduct extensive experimental evaluations across successive training stages, dividing the benchmarks into two primary categories: (i) discriminative tasks (classification and retrieval) and (ii) generative multimodal understanding. For discriminative tasks, we evaluate zero-shot classification performance on ImageNet-1k \cite{deng2009imagenet}, ImageNet-v2 \cite{recht2019imagenet}, and ImageNet-R \cite{beyer2020we}. Cross-modal evaluations include text-to-image and image-to-text retrieval tasks on five datasets, including COCO \cite{chen2015microsoft}, Flickr30k \cite{plummer2015flickr30k}, and the more recent DCI \cite{urbanek2024picture}, IIW \cite{garg2024imageinwords}, and Urban-1k \cite{zhang2024long}.
Regarding multimodal understanding, the evaluation protocols span four distinct sub-domains. (1) General Visual Question Answering (VQA): MMBench \cite{liu2024mmbench}, MMStar \cite{chen2024we}, and HallusionBench \cite{guan2024hallusionbench}. (2) Multimodal Reasoning: MMMU \cite{yue2024mmmu}, MathVista \cite{lu2023mathvista}, MMVet \cite{yu2023mm}, SEEDBench \cite{li2023seed}, and ScienceQA \cite{lu2022learn}. (3) OCR \& Chart: AI2D \cite{kembhavi2016diagram}, OCRBench \cite{liu2024ocrbench}, DocVQA \cite{mathew2021docvqa}, ChartQA \cite{masry2022chartqa}, InfoVQA \cite{mathew2022infographicvqa}, and TextVQA \cite{singh2019towards}. (4) Grounding \& Counting: BLINK \cite{fu2024blink}, CountBenchQA \cite{beyer2024paligemma}, PointBench \cite{cheng2025pointarenaprobingmultimodalgrounding}, the RefCOCO series \cite{yu2016modeling}, VisualWebBench \cite{liu2024visualwebbench}, and FSC-147 \cite{ranjan2021learning}. OCRBench scores are normalized to a \texttt{[0, 100]} scale to ensure consistency during averaging. For the FSC-147 counting task, scores are integrated into the final average as $100 - \text{FSC-147}_{val}$, with individual errors capped at $100$. Results for RefCOCO series represent the mean performance over the validation and test splits of each dataset. To ensure fair comparison, \textit{all multimodal understanding tasks are benchmarked using the standard VLMEvalKit~\cite{duan2024vlmevalkit} framework.}

\subsection{Main Results}


\begin{table*}[t]
    \centering
    \caption{Comparison on Zero-shot Classification and Retrieval benchmarks.}
    \label{tab:zeroshot_classification_retrieval}
    \resizebox{1.0\textwidth}{!}{%
    \setlength{\tabcolsep}{3pt} 
    \begin{tabular}{l c ccc cccc}
        \toprule
        \multirow{2}{*}{\textbf{Model}} & \multirow{2}{*}{\textbf{Params}} & \multicolumn{3}{c}{\textbf{Zero-shot Classification}} & \multicolumn{4}{c}{\textbf{Zero-shot Retrieval}} \\
        \cmidrule(lr){3-5} \cmidrule(lr){6-9}
        & & IN-1k val & IN v2 & IN REAL & COCO T2I & COCO I2T & FLICKR T2I & FLICKR I2T \\
        \midrule
        SigLIP-so400m/14~\cite{zhai2023sigmoid} & 0.4B & 83.2 & 77.1 & 87.5 & 52.0 & 70.2 & 80.5 & 93.5 \\
        SigLIP2-so400m/14~\cite{tschannen2025siglip} & 0.4B & 84.1 & 78.7 & 88.1 & 55.8 & 71.7 & 85.7 & 94.9 \\
        SigLIP2-g/16~\cite{tschannen2025siglip} & 1B & \textbf{85.0} & \textbf{79.8} & 88.5 & 56.1 & 72.8 & \textbf{86.0} & 95.4 \\
        Seed-ViT/14~\cite{guo2025seed1} & 0.5B & 83.6 & 77.6 & - & - & - & - & - \\
        \midrule
        \rowcolor{gray!10} \textbf{FineViT/14} & 0.86B & 84.2 & 75.5 & \textbf{88.7} & \textbf{60.7} & \textbf{80.7} & 84.8 & \textbf{96.7} \\
        \bottomrule
    \end{tabular}}
\end{table*}

\begin{table*}[t]
    \centering
    \caption{Comparison on Long text Zero-shot Retrieval benchmarks.}
    \label{tab:long_text_retrieval}
    \resizebox{1.0\textwidth}{!}{%
    \setlength{\tabcolsep}{5pt} 
        \begin{tabular}{l c cc cc cc}
            \toprule
            \multirow{2}{*}{\textbf{Model}} & \multirow{2}{*}{\textbf{Params}} & \multicolumn{6}{c}{\textbf{Long-text Zero-shot Retrieval}} \\
            \cmidrule(lr){3-8}
             & & DCI T2I & DCI I2T & IIW T2I & IIW I2T & Urban-1k T2I & Urban-1k I2T \\
            \midrule
            SigLIP-so400m/14~\cite{zhai2023sigmoid} & 0.4B & 61.5 & 63.3 & 91.8 & 92.5 & 73.8 & 74.7 \\
            SigLIP2-so400m/14~\cite{tschannen2025siglip} & 0.4B & 66.8 & 66.9 & 95.6 & 94.8 & 75.6 & 78.1 \\
            LongCLIP-L/14~\cite{zhang2024long}  & 0.3B & 63.8 & 56.2 & 96.1 & 94.1 & 84.0 & 81.1 \\
            FixCLIP-L/14~\cite{wang2025fix} & 0.3B & 74.2 & 72.0 & 98.2 & 97.9 & 96.3 & 93.7 \\
            \midrule
            \rowcolor{gray!10} \textbf{FineViT/14} & 0.86B & \textbf{84.8} & \textbf{83.4} & \textbf{99.7} & \textbf{99.8} & \textbf{99.1} & \textbf{98.9} \\
            \bottomrule
        \end{tabular}%
    }
\end{table*}

As shown in Table \ref{tab:zeroshot_classification_retrieval}, we compare FineViT against state-of-the-art vision language models, including SigLIP, SigLIP2, and Seed-ViT. Despite having fewer parameters than the SigLIP2-g (1B) variant, FineViT achieves competitive or superior performance across most metrics. Specifically, on the ImageNet-1k validation set, FineViT attains a top-1 accuracy of $84.2\%$, surpassing SigLIP-so400m ($83.2\%$) and Seed-ViT ($83.6\%$). FineViT demonstrates remarkable improvements on retrieval benchmarks. On MS COCO, FineViT achieves a text-to-image (T2I) score of $60.7\%$ and an image-to-text (I2T) score of $80.7\%$, significantly outperforming SigLIP2-so400m by margins of $4.9\%$ and $9.0\%$, respectively. Similar trends are observed on Flickr30K, where FineViT sets a new state-of-the-art of $96.7\%$ in I2T retrieval.
\noindent\textbf{Zero-shot Classification and Retrieval.} 

\noindent\textbf{Long text Zero-shot Retrieval.} 
Benefiting from dense recaptions, FineViT supports long text retrieval. To evaluate its capability in handling complex and dense textual descriptions, we benchmark FineViT on three long text retrieval datasets: DCI, IIW, and Urban-1K. The results, summarized in Table \ref{tab:long_text_retrieval}, highlight the exceptional fine-grained understanding capabilities of our method. Compared to baselines specifically optimized for text alignment, such as FixCLIP, FineViT achieves substantial gains. Notably, on the challenging DCI dataset, our model reaches $84.8\%$ in T2I retrieval, exceeding FixCLIP ($74.2\%$) by over $10$ points and the SigLIP2 baseline ($66.8\%$) by $18$ points. On the Urban-1K benchmark, FineViT approaches saturation with near perfect scores ($99.1\%$ T2I/$98.9\%$ I2T), demonstrating robust alignment between visual features and detailed textual captions. Several visual comparisons are shown in Figure~\ref{fig:longretrieval}. \textit{Additional visualization results are provided in the appendix.}

\begin{figure}[t]
  \centering
  \includegraphics[width=1.0\columnwidth]{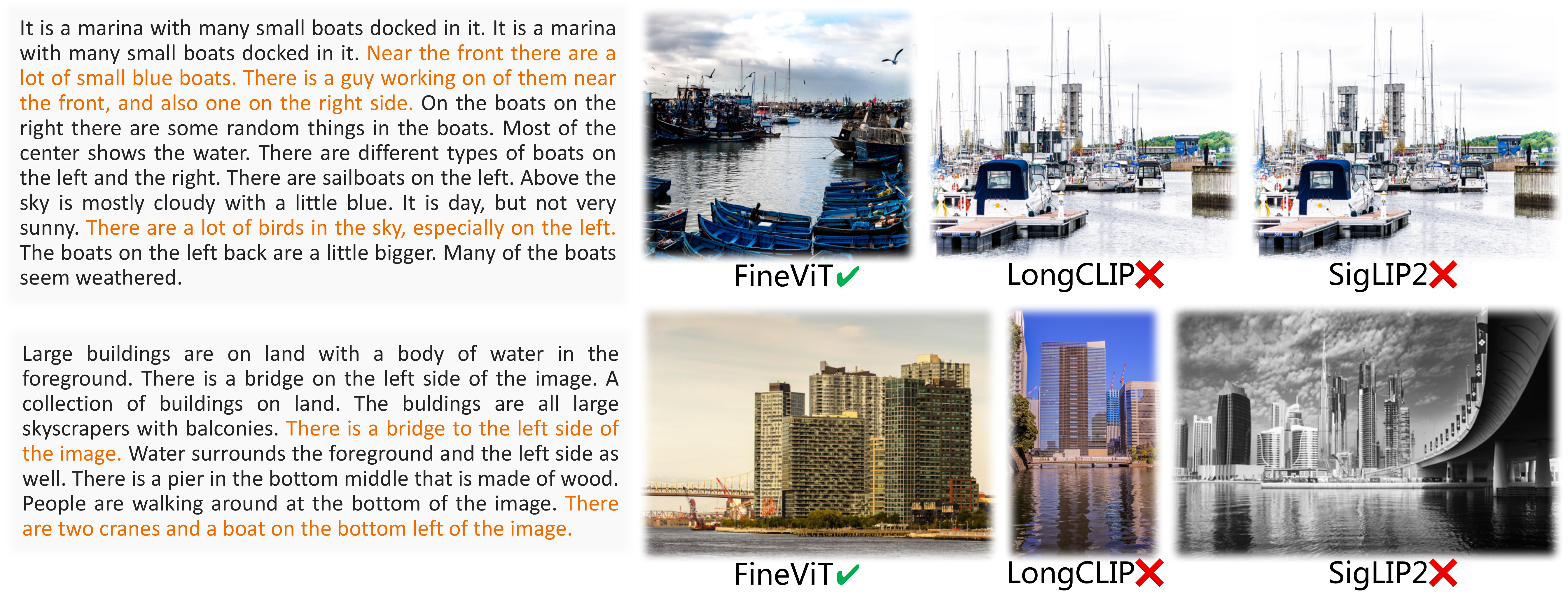}
  \caption{Illustration on long text zero-shot retrieval. Distinctive attributes within the long caption are highlighted in {\color{orange}{orange}} to help identify the correct image.}
  \label{fig:longretrieval}
\end{figure}

\begin{table*}[!t]
    \centering
    \caption{Performance comparison across vision language models on various multimodal understanding benchmarks. Aquila-VL uses Qwen2.5 1.5B LLM, all others utilize Qwen3 1.7B LLM as language backbone. \textbf{All results are evaluated using the standard VLMEvalKit~\cite{duan2024vlmevalkit} framework and identical APIs.}}
    \setlength\tabcolsep{6pt}
    \Large
    \resizebox{1.0\textwidth}{!}{\begin{tabular}{ccc|ccccc}
        \toprule
         &\bf{Task}&\bf{Benchmark}&\bf{FineViT-VL}&\bf{Qwen3-VL~\cite{Qwen3-VL}}&\bf{Intern3.5-VL~\cite{wang2025internvl3}}&\bf{Aquila-VL~\cite{gu2024infinity}}\\
         \midrule
         \multirow{8}{*}{General}&\multirow{3}{*}{\makecell{General \\ VQA}}&MMBench$_{dev\_en\_v11}$&\textbf{79.33}&76.47&76.08&75.69\\
         &&MMStar&\textbf{60.87}&56.60&57.33&54.53\\
         &&HallusionBench&46.54&\textbf{51.89}&48.18&42.07\\
         \cmidrule(l){2-7} 
         &\multirow{5}{*}{\makecell{Multimodal \\ Reasoning}}&MMMU$_{val}$&44.33&44.11&\textbf{49.33}&43.77\\
         &&MathVista$_{mini}$&\textbf{67.70}&55.10&61.50&59.30\\
         &&MMVet&55.05&60.18&\textbf{65.20}&40.22\\
         &&SEEDBench$_{img}$&\textbf{77.44}&74.78&75.31&73.90\\
         &&ScienceQA$_{test}$&\textbf{95.93}&86.47&91.17&95.33\\
         \midrule
         \rowcolor{gray!10} \multicolumn{3}{c|}{Average}&{\textbf{65.90}}&{63.20}&65.51&60.60\\
         \midrule
         \multirow{14}{*}{Local}&\multirow{6}{*}{\makecell{OCR \\\& Chart}}&AI2D$_{test}$&\textbf{82.25}&76.46&79.01&75.06\\    
         &&OCRBench&841&\textbf{869}&835&772\\
         &&DocVQA$_{test}$&93.41&\textbf{93.47}&89.49&74.29\\
         &&ChartQA$_{test}$&\textbf{85.60}&77.72&80.76&76.52\\
         &&InfoVQA$_{test}$&\textbf{72.20}&71.98&70.79&46.67\\
         &&TextVQA$_{val}$&\textbf{83.26}&80.77&76.72&76.27\\
         \cmidrule(l){2-7} 
         &\multirow{8}{*}{\makecell{Grounding \\\& Counting}}&BLINK$_{count}$
&\textbf{67.50}&63.33&65.00&66.67\\
         &&CountBenchQA&\textbf{93.84}&86.65&70.22&80.69\\
         &&PointBench&\textbf{70.21}&64.89&60.63&63.29\\
         &&RefCOCO$_{avg}$&\textbf{90.60}&88.04&88.28&63.51\\
         &&RefCOCO{+}$_{avg}$&\textbf{85.78}&80.07&82.49&57.69\\
         &&RefCOCOg$_{avg}$&\textbf{88.81}&85.43&85.28&58.71\\
         &&VisualWebBench&\textbf{61.26}&60.24&44.85&41.43\\
         &&100-FSC-147$_{val}$&\textbf{81.39}&76.27&77.50&70.43\\
         \midrule
         \rowcolor{gray!10} \multicolumn{3}{c|}{Average}&\textbf{81.44}&78.02&75.32&66.32\\
         \bottomrule
    \end{tabular}}
    \label{tab:main_vlm}
\end{table*}

\noindent\textbf{Multimodal Understanding Tasks.} 
To evaluate the efficacy of FineViT-VL, we conduct a comprehensive comparison against leading small scale VLMs, including Qwen3-VL, Intern3.5-VL, and Aquila-VL. As shown in Table \ref{tab:main_vlm}, with a language model size of approximately $1.7$B, FineViT yields competitive results across the General QA and Multimodal Reasoning benchmarks. Notebly, FineViT-VL achieves substantial performance gain in localized tasks, outperforming the second-best competitor by an average margin of $3.42\%$. For OCR and Chart Understanding, FineViT-VL exhibits marked advantages on the AI2D and ChartQA benchmarks, demonstrating its robust capability in parsing complex visual structures and dense textual information. Beyond text centric understanding, FineViT-VL also demonstrates superior spatial awareness in Grounding and Counting benchmarks, such as CountBenchQA and the RefCOCO series. We attribute this local precision to the large-scale, multi-granularity alignment training enabled by FineCap-450M. This approach captures fine visual details and precise instance level dependencies that are often overlooked by standard global alignment methods.


\subsection{Ablation Study}
\begin{table*}[!t]
    \centering
    \caption{Benchmarking multimodal performance of vision encoders and LLMs}
    \setlength\tabcolsep{6pt}
    \resizebox{1.0\textwidth}{!}{\begin{tabular}{ccc|cc|cc}
        \toprule
         &\multirow{2}{*}{\bf{Task}}&\multirow{2}{*}{\bf{Benchmark}}&\multicolumn{2}{c|}{\textbf{LLM: Qwen3-1.7B}}&\multicolumn{2}{c}{\textbf{LLM: Qwen3-8B}}\\
         &&&FineViT&SigLIP2-naflex&FineViT&SigLIP2-naflex\\
         \midrule
         \multirow{8}{*}{General}&\multirow{3}{*}{\makecell{General \\ VQA}}&MMBench$_{dev\_en\_v11}$&{66.25}&{66.87}&{74.54}&{76.32}\\
         &&MMStar&{44.60}&{44.73}&{55.93}&{56.60}\\
         &&HallusionBench&{30.04}&{30.74}&{39.83}&{39.08}\\
         \cmidrule(l){2-7} 
         &\multirow{5}{*}{\makecell{Multimodal \\ Reasoning}}&MMMU$_{val}$&{41.22}&{41.44}&{45.89}&{48.56}\\
         &&MathVista$_{mini}$&{49.00}&{43.20}&{58.80}&{60.50}\\
         &&MMVet&{32.66}&{28.90}&{43.12}&{40.41}\\
         &&SEEDBench$_{img}$&{72.33}&{71.52}&{75.53}&{74.76}\\
         &&ScienceQA$_{test}$&{78.14}&{80.07}&{84.58}&{86.17}\\
         \midrule
         \rowcolor{gray!10} \multicolumn{3}{c|}{Average}&{\textbf{51.78}}&{50.93}&{59.78}&{\textbf{60.30}}\\
         \midrule
         \multirow{14}{*}{Local}&\multirow{6}{*}{\makecell{OCR \\\& Chart}}&AI2D$_{test}$&{70.85}&{68.43}&{77.23}&{77.49}\\  
         &&OCRBench&{704}&{533}&{698}&{550}\\
         &&DocVQA$_{test}$&{83.49}&{64.07}&{83.84}&{73.05}\\
         &&ChartQA$_{test}$&{73.00}&{63.48}&{75.20}&{75.16}\\
         &&InfoVQA$_{test}$&{52.40}&{39.60}&{54.52}&{48.55}\\
         &&TextVQA$_{val}$&{75.65}&{67.74}&{77.91}&{72.26}\\
         \cmidrule(l){2-7} 
         &\multirow{8}{*}{\makecell{Grounding \\\& Counting}}&BLINK$_{count}$
&{57.50}&{51.67}&{72.50}&{62.50}\\
         &&CountBenchQA&{83.37}&{77.82}&{85.01}&{71.66}\\
         &&PointBench&{65.43}&{59.04}&{70.21}&{63.83}\\
         &&RefCOCO$_{avg}$&{78.93}&{67.61}&{85.53}&{81.92}\\
         &&RefCOCO{+}$_{avg}$&{68.00}&{57.34}&{76.35}&{74.32}\\
         &&RefCOCOg$_{avg}$&{76.67}&{64.16}&{82.78}&{80.06}\\
         &&VisualWebBench&{57.11}&{41.29}&{58.92}&{50.75}\\
         &&100-FSC-147$_{val}$&{68.20}&{47.10}&{59.02}&{53.57}\\
         \midrule
         \rowcolor{gray!10} \multicolumn{3}{c|}{Average}&{\textbf{70.07}}&{58.76}&{\textbf{73.49}}&{67.15}\\
         \bottomrule
    \end{tabular}}
    \label{tab:ablation_vit_llm}
\end{table*}
\noindent\textbf{Comparison of Vision Encoders across Various LLM Scales.}
Table \ref{tab:ablation_vit_llm} presents a comprehensive quantitative evaluation of FineViT against the SigLIP2-naflex baseline across diverse multimodal benchmarks. We instantiate these vision encoders with various LLMs to verify their scalability and compatibility.
While both encoders exhibit comparable performance on global level VQA tasks (\textit{e.g.}, MMBench, MMStar), FineViT demonstrates a decisive advantage in local perception tasks. Specifically, when paired with Qwen3-1.7B, FineViT achieves a substantial average margin of $11.31\%$ over SigLIP2 in local centric benchmarks. This performance gap is most pronounced in OCR and document understanding, for instance, FineViT yields an absolute improvement of $18.52\%$ on DocVQA and $13.68\%$ on InfoVQA. These results suggest that FineViT better preserves the high resolution spatial features essential for textual and structural recognition.
The performance gain persists across different LLM capacities. Transitioning from the $1.7$B to the $8$B Qwen3 variant, FineViT consistently outperforms SigLIP2 in grounding and counting tasks, such as BLINK-Count ($+10.0\%$) and CountBenchQA ($+13.35\%$). Notably, FineViT shows superior sample efficiency in grounding, as evidenced by its consistent lead across all RefCOCO variants. This indicates that our architectural refinements facilitate a more precise alignment between visual tokens and linguistic queries, thereby mitigating the spatial ambiguity often encountered in complex multimodal reasoning.

\begin{table*}[!t]
    \centering
    \caption{Multimodal performance of FineViT across different training stages.}
    \setlength\tabcolsep{8pt}
    \resizebox{1.0\columnwidth}{!}{\begin{tabular}{ccc|ccc}
        \toprule
         &\bf{Task}&\bf{Benchmark}&\bf{Stage I: MIM}&\bf{Stage II: CL}&\bf{Stage III: Align}\\
         \midrule
         \multirow{8}{*}{General}&\multirow{3}{*}{\makecell{General \\ VQA}}&MMBench$_{dev\_en\_v11}$&{62.15}&{67.18}&{69.97}\\
         &&MMStar&{43.73}&{46.00}&{49.20}\\
         &&HallusionBench&{27.48}&{32.45}&{34.13}\\
         \cmidrule(l){2-6} 
         &\multirow{5}{*}{\makecell{Multimodal \\ Reasoning}}&MMMU$_{val}$&{37.44}&{40.33}&{41.44}\\
         &&MathVista$_{mini}$&{44.60}&{48.20}&{52.70}\\
         &&MMVet&{27.80}&{36.65}&{35.83}\\
         &&SEEDBench$_{img}$&{71.77}&{73.31}&{74.34}\\
         &&ScienceQA$_{test}$&{75.21}&{80.52}&{81.01}\\
         \midrule
         \rowcolor{gray!10} \multicolumn{3}{c|}{Average}&{48.77}&{53.08}&{\textbf{54.83}}\\
         \midrule
         \multirow{14}{*}{Local}&\multirow{6}{*}{\makecell{OCR \\\& Chart}}&AI2D$_{test}$&{65.90}&{69.11}&{71.70}\\   
         &&OCRBench&{417}&{681}&{745}\\
         &&DocVQA$_{test}$&{64.26}&{79.85}&{85.33}\\
         &&ChartQA$_{test}$&{67.16}&{73.00}&{77.56}\\
         &&InfoVQA$_{test}$&{35.18}&{48.70}&{56.25}\\
         &&TextVQA$_{val}$&{50.84}&{73.82}&{77.35}\\
         \cmidrule(l){2-6} 
         &\multirow{8}{*}{\makecell{Grounding \\\& Counting}}&BLINK$_{count}$
&{55.00}&{56.67}&{60.00}\\
         &&CountBenchQA&{56.47}&{72.07}&{90.76}\\
         &&PointBench&{54.79}&{59.04}&{67.55}\\
         &&RefCOCO$_{avg}$&{77.93}&{72.93}&{83.70}\\
         &&RefCOCO{+}$_{avg}$&{66.47}&{63.18}&{73.87}\\
         &&RefCOCOg$_{avg}$&{76.59}&{69.51}&{80.92}\\
         &&VisualWebBench&{39.98}&{53.03}&{56.21}\\
         &&100$-$FSC-147$_{val}$&{48.04}&{48.57}&{70.33}\\
         \midrule
         \rowcolor{gray!10} \multicolumn{3}{c|}{Average}&{57.12}&{64.80}&{\textbf{73.29}}\\
         \bottomrule
    \end{tabular}}
    \label{tab:comparison_different_stages}
\end{table*}

\noindent\textbf{Effectiveness of Progressive Training Stages.} 
We systematically evaluate the contribution of each training phase in Table \ref{tab:comparison_different_stages}. While Stage I (MIM) constructs a robust foundational representation, it lacks the high-level semantic discriminability required for complex reasoning. The subsequent Stage II (contrastive learning) bridges this gap, yielding substantial gains in general VQA. However, the global nature of contrastive objectives occasionally suppresses instance-level spatial sensitivity, as evidenced by performance fluctuations in grounding tasks (\textit{e.g.}, RefCOCO). Crucially, the introduction of Stage III, enriched by our large scale FineCap-450M dataset, effectively enhancing the spatial perceptions. This improvement is most prominent in spatially dense benchmarks: the OCRBench score surges from $681$ to $745$, and CountBenchQA exhibits a remarkable $18.69\%$ improvement. These results demonstrate that the synergy between our multi-stage pipeline and the FineCap-450M corpus not only alleviates the localization bottleneck but also empowers FineViT with superior granularity in document parsing and visual grounding.

\begin{figure}[t]
  \centering
  \includegraphics[width=1.0\columnwidth]{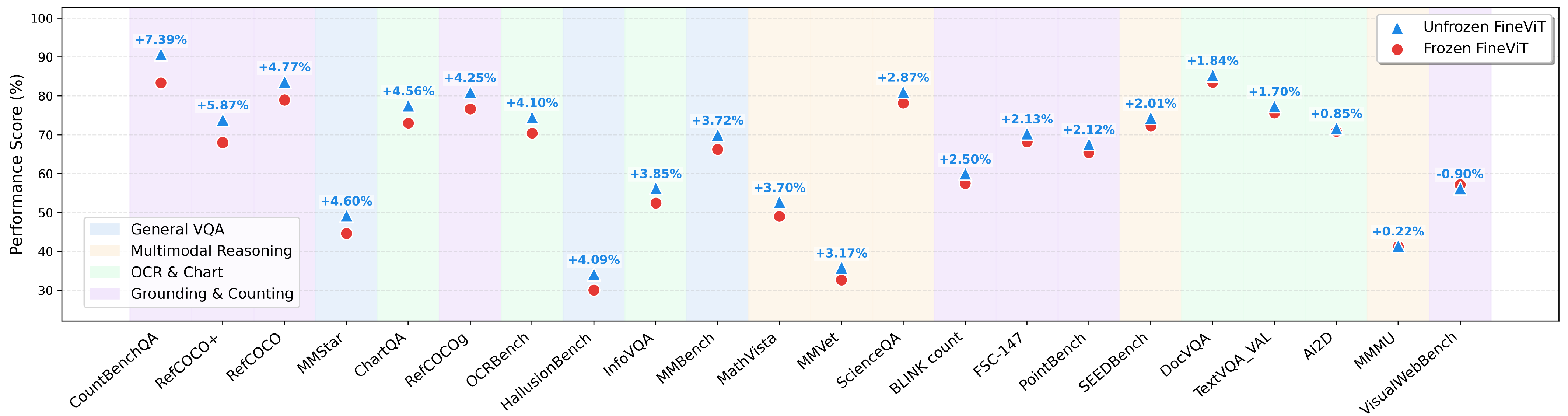}
  \caption{Performance distribution: unfrozen vs frozen FineViT.}
  \label{fig:ablation_frozen_vs_unfrozen}
\end{figure}

\noindent\textbf{Frozen vs. Unfrozen Visual Backbone.} 
To investigate the contribution of visual feature adaption, we compare the performance of frozen vs. unfrozen FineViT. The results, categorized by task type and sorted by relative gain, are illustrated in Figure~\ref{fig:ablation_frozen_vs_unfrozen}.
Unfreezing the vision encoder yields the most significant gains in \textit{Local} tasks (\textit{e.g.}, RefCOCO, ChartQA). While a frozen encoder suffices for general tasks, updating the visual backbone is indispensable for benchmarks demanding fine-grained visual features and exact positioning.

\section{Conclusion}
In this paper, we introduced FineViT, a state-of-the-art vision encoder trained from scratch to address the limitations of coarse-grained perception in existing visual foundation models. To achieve precise visual understanding, we propose a progressive training paradigm driven by our curated FineCap-450M dataset, the largest fine-grained annotated dataset to date. FineViT effectively captures intricate spatial relationships and dense visual details. Extensive evaluations demonstrate that FineViT matches the zero-shot performance of leading models like SigLIP2, while uniquely excelling in long-context retrieval. Furthermore, when integrated into MLLMs, FineViT consistently outperforms contemporary architectures such as SigLIP2 and Qwen-ViT. We would release FineViT and the FineCap-450M dataset to advance research in fine-grained visual perception. \textbf{Limitation.} FineViT is currently restricted to static images, leaving its potential for video-based temporal reasoning to be explored in future work.

\clearpage
\bibliographystyle{unsrt}
\bibliography{references}  

@String(CVPR= {IEEE Conf. Comput. Vis. Pattern Recog.})

@String(ECCV= {Eur. Conf. Comput. Vis.})

@String(CVPR  = {CVPR})

@String(ECCV  = {ECCV})

@inproceedings{radford2021learning,
  title={Learning transferable visual models from natural language supervision},
  author={Radford, Alec and Kim, Jong Wook and Hallacy, Chris and Ramesh, Aditya and Goh, Gabriel and Agarwal, Sandhini and Sastry, Girish and Askell, Amanda and Mishkin, Pamela and Clark, Jack and others},
  booktitle={International conference on machine learning},
  pages={8748--8763},
  year={2021},
  organization={PmLR}
}

@inproceedings{zhai2023sigmoid,
  title={Sigmoid loss for language image pre-training},
  author={Zhai, Xiaohua and Mustafa, Basil and Kolesnikov, Alexander and Beyer, Lucas},
  booktitle={Proceedings of the IEEE/CVF international conference on computer vision},
  pages={11975--11986},
  year={2023}
}

@article{tschannen2025siglip,
  title={Siglip 2: Multilingual vision-language encoders with improved semantic understanding, localization, and dense features},
  author={Tschannen, Michael and Gritsenko, Alexey and Wang, Xiao and Naeem, Muhammad Ferjad and Alabdulmohsin, Ibrahim and Parthasarathy, Nikhil and Evans, Talfan and Beyer, Lucas and Xia, Ye and Mustafa, Basil and others},
  journal={arXiv preprint arXiv:2502.14786},
  year={2025}
}

@article{sun2023eva,
  title={Eva-clip: Improved training techniques for clip at scale},
  author={Sun, Quan and Fang, Yuxin and Wu, Ledell and Wang, Xinlong and Cao, Yue},
  journal={arXiv preprint arXiv:2303.15389},
  year={2023}
}

@article{sun2024eva,
  title={Eva-clip-18b: Scaling clip to 18 billion parameters},
  author={Sun, Quan and Wang, Jinsheng and Yu, Qiying and Cui, Yufeng and Zhang, Fan and Zhang, Xiaosong and Wang, Xinlong},
  journal={arXiv preprint arXiv:2402.04252},
  year={2024}
}

@inproceedings{he2020momentum,
  title={Momentum contrast for unsupervised visual representation learning},
  author={He, Kaiming and Fan, Haoqi and Wu, Yuxin and Xie, Saining and Girshick, Ross},
  booktitle={Proceedings of the IEEE/CVF conference on computer vision and pattern recognition},
  pages={9729--9738},
  year={2020}
}

@article{wu2025qwen,
  title={Qwen-image technical report},
  author={Wu, Chenfei and Li, Jiahao and Zhou, Jingren and Lin, Junyang and Gao, Kaiyuan and Yan, Kun and Yin, Sheng-ming and Bai, Shuai and Xu, Xiao and Chen, Yilei and others},
  journal={arXiv preprint arXiv:2508.02324},
  year={2025}
}

@inproceedings{shi2025scaling,
  title={Scaling vision pre-training to 4k resolution},
  author={Shi, Baifeng and Li, Boyi and Cai, Han and Lu, Yao and Liu, Sifei and Pavone, Marco and Kautz, Jan and Han, Song and Darrell, Trevor and Molchanov, Pavlo and others},
  booktitle={Proceedings of the IEEE/CVF Conference on Computer Vision and Pattern Recognition},
  pages={9631--9640},
  year={2025}
}

@article{yao2024minicpm,
  title={Minicpm-v: A gpt-4v level mllm on your phone},
  author={Yao, Yuan and Yu, Tianyu and Zhang, Ao and Wang, Chongyi and Cui, Junbo and Zhu, Hongji and Cai, Tianchi and Li, Haoyu and Zhao, Weilin and He, Zhihui and others},
  journal={arXiv preprint arXiv:2408.01800},
  year={2024}
}

@article{wiedmann2025finevision,
  title={Finevision: Open data is all you need},
  author={Wiedmann, Luis and Zohar, Orr and Mahla, Amir and Wang, Xiaohan and Li, Rui and Frere, Thibaud and von Werra, Leandro and Gosthipaty, Aritra Roy and Marafioti, Andr{\'e}s},
  journal={arXiv preprint arXiv:2510.17269},
  year={2025}
}

@article{cui2025paddleocr,
  title={Paddleocr 3.0 technical report},
  author={Cui, Cheng and Sun, Ting and Lin, Manhui and Gao, Tingquan and Zhang, Yubo and Liu, Jiaxuan and Wang, Xueqing and Zhang, Zelun and Zhou, Changda and Liu, Hongen and others},
  journal={arXiv preprint arXiv:2507.05595},
  year={2025}
}

@inproceedings{liu2024grounding,
  title={Grounding dino: Marrying dino with grounded pre-training for open-set object detection},
  author={Liu, Shilong and Zeng, Zhaoyang and Ren, Tianhe and Li, Feng and Zhang, Hao and Yang, Jie and Jiang, Qing and Li, Chunyuan and Yang, Jianwei and Su, Hang and others},
  booktitle={European conference on computer vision},
  pages={38--55},
  year={2024},
  organization={Springer}
}

@inproceedings{zheng2025granvit,
  title={GranViT: A Fine-Grained Vision Model With Autoregressive Perception For MLLMs},
  author={Zheng, Guanghao and Shi, Bowen and Xu, Mingxing and Sun, Ruoyu and Zhao, Peisen and Dai, Wenrui and Zou, Junni and Xiong, Hongkai and ZHANG, XIAOPENG and Tian, Qi and others},
  booktitle={The Fourteenth International Conference on Learning Representations},
  year={2025}
}

@article{li2024if,
  title={What if we recaption billions of web images with llama-3?},
  author={Li, Xianhang and Tu, Haoqin and Hui, Mude and Wang, Zeyu and Zhao, Bingchen and Xiao, Junfei and Ren, Sucheng and Mei, Jieru and Liu, Qing and Zheng, Huangjie and others},
  journal={arXiv preprint arXiv:2406.08478},
  year={2024}
}

@inproceedings{chen2024internvl,
  title={Internvl: Scaling up vision foundation models and aligning for generic visual-linguistic tasks},
  author={Chen, Zhe and Wu, Jiannan and Wang, Wenhai and Su, Weijie and Chen, Guo and Xing, Sen and Zhong, Muyan and Zhang, Qinglong and Zhu, Xizhou and Lu, Lewei and others},
  booktitle={Proceedings of the IEEE/CVF conference on computer vision and pattern recognition},
  pages={24185--24198},
  year={2024}
}

@article{chen2024expanding,
  title={Expanding performance boundaries of open-source multimodal models with model, data, and test-time scaling},
  author={Chen, Zhe and Wang, Weiyun and Cao, Yue and Liu, Yangzhou and Gao, Zhangwei and Cui, Erfei and Zhu, Jinguo and Ye, Shenglong and Tian, Hao and Liu, Zhaoyang and others},
  journal={arXiv preprint arXiv:2412.05271},
  year={2024}
}

@article{zhu2025internvl3,
  title={Internvl3: Exploring advanced training and test-time recipes for open-source multimodal models},
  author={Zhu, Jinguo and Wang, Weiyun and Chen, Zhe and Liu, Zhaoyang and Ye, Shenglong and Gu, Lixin and Tian, Hao and Duan, Yuchen and Su, Weijie and Shao, Jie and others},
  journal={arXiv preprint arXiv:2504.10479},
  year={2025}
}

@article{bai2025qwen2,
  title={Qwen2. 5-vl technical report},
  author={Bai, Shuai and Chen, Keqin and Liu, Xuejing and Wang, Jialin and Ge, Wenbin and Song, Sibo and Dang, Kai and Wang, Peng and Wang, Shijie and Tang, Jun and others},
  journal={arXiv preprint arXiv:2502.13923},
  year={2025}
}

@article{guo2025seed1,
  title={Seed1. 5-vl technical report},
  author={Guo, Dong and Wu, Faming and Zhu, Feida and Leng, Fuxing and Shi, Guang and Chen, Haobin and Fan, Haoqi and Wang, Jian and Jiang, Jianyu and Wang, Jiawei and others},
  journal={arXiv preprint arXiv:2505.07062},
  year={2025}
}

@inproceedings{fini2025multimodal,
  title={Multimodal autoregressive pre-training of large vision encoders},
  author={Fini, Enrico and Shukor, Mustafa and Li, Xiujun and Dufter, Philipp and Klein, Michal and Haldimann, David and Aitharaju, Sai and da Costa, Victor G Turrisi and B{\'e}thune, Louis and Gan, Zhe and others},
  booktitle={Proceedings of the Computer Vision and Pattern Recognition Conference},
  pages={9641--9654},
  year={2025}
}

@misc{liu2024llavanext,
    title={LLaVA-NeXT: Improved reasoning, OCR, and world knowledge},
    url={https://llava-vl.github.io/blog/2024-01-30-llava-next/},
    author={Liu, Haotian and Li, Chunyuan and Li, Yuheng and Li, Bo and Zhang, Yuanhan and Shen, Sheng and Lee, Yong Jae},
    month={January},
    year={2024}
}

@article{schuhmann2021laion,
  title={Laion-400m: Open dataset of clip-filtered 400 million image-text pairs},
  author={Schuhmann, Christoph and Vencu, Richard and Beaumont, Romain and Kaczmarczyk, Robert and Mullis, Clayton and Katta, Aarush and Coombes, Theo and Jitsev, Jenia and Komatsuzaki, Aran},
  journal={arXiv preprint arXiv:2111.02114},
  year={2021}
}

@inproceedings{LLaVA-OneVision-1.5,
  title={LLaVA-OneVision-1.5: Fully Open Framework for Democratized Multimodal Training},
  author={An, Xiang and Xie, Yin and Yang, Kaicheng and Zhang, Wenkang and Zhao, Xiuwei and Cheng, Zheng and Wang, Yirui and Xu, Songcen and Chen, Changrui and Wu, Chunsheng and Tan, Huajie and Li, Chunyuan and Yang, Jing and Yu, Jie and Wang, Xiyao and Qin, Bin and Wang, Yumeng and Yan, Zizhen and Feng, Ziyong and Liu, Ziwei and Li, Bo and Deng, Jiankang},
  booktitle={arXiv},  
  year={2025}
 }

@article{chen2024we,
  title={Are we on the right way for evaluating large vision-language models?},
  author={Chen, Lin and Li, Jinsong and Dong, Xiaoyi and Zhang, Pan and Zang, Yuhang and Chen, Zehui and Duan, Haodong and Wang, Jiaqi and Qiao, Yu and Lin, Dahua and others},
  journal={Advances in Neural Information Processing Systems},
  volume={37},
  pages={27056--27087},
  year={2024}
}

@inproceedings{liu2024mmbench,
  title={Mmbench: Is your multi-modal model an all-around player?},
  author={Liu, Yuan and Duan, Haodong and Zhang, Yuanhan and Li, Bo and Zhang, Songyang and Zhao, Wangbo and Yuan, Yike and Wang, Jiaqi and He, Conghui and Liu, Ziwei and others},
  booktitle={European conference on computer vision},
  pages={216--233},
  year={2024},
  organization={Springer}
}

@inproceedings{guan2024hallusionbench,
  title={Hallusionbench: an advanced diagnostic suite for entangled language hallucination and visual illusion in large vision-language models},
  author={Guan, Tianrui and Liu, Fuxiao and Wu, Xiyang and Xian, Ruiqi and Li, Zongxia and Liu, Xiaoyu and Wang, Xijun and Chen, Lichang and Huang, Furong and Yacoob, Yaser and others},
  booktitle={Proceedings of the IEEE/CVF Conference on Computer Vision and Pattern Recognition},
  pages={14375--14385},
  year={2024}
}

@article{li2023seed,
  title={Seed-bench: Benchmarking multimodal llms with generative comprehension},
  author={Li, Bohao and Wang, Rui and Wang, Guangzhi and Ge, Yuying and Ge, Yixiao and Shan, Ying},
  journal={arXiv preprint arXiv:2307.16125},
  year={2023}
}

@inproceedings{yue2024mmmu,
  title={Mmmu: A massive multi-discipline multimodal understanding and reasoning benchmark for expert agi},
  author={Yue, Xiang and Ni, Yuansheng and Zhang, Kai and Zheng, Tianyu and Liu, Ruoqi and Zhang, Ge and Stevens, Samuel and Jiang, Dongfu and Ren, Weiming and Sun, Yuxuan and others},
  booktitle={Proceedings of the IEEE/CVF Conference on Computer Vision and Pattern Recognition},
  pages={9556--9567},
  year={2024}
}

@article{lu2023mathvista,
  title={Mathvista: Evaluating mathematical reasoning of foundation models in visual contexts},
  author={Lu, Pan and Bansal, Hritik and Xia, Tony and Liu, Jiacheng and Li, Chunyuan and Hajishirzi, Hannaneh and Cheng, Hao and Chang, Kai-Wei and Galley, Michel and Gao, Jianfeng},
  journal={arXiv preprint arXiv:2310.02255},
  year={2023}
}

@article{yu2023mm,
  title={Mm-vet: Evaluating large multimodal models for integrated capabilities},
  author={Yu, Weihao and Yang, Zhengyuan and Li, Linjie and Wang, Jianfeng and Lin, Kevin and Liu, Zicheng and Wang, Xinchao and Wang, Lijuan},
  journal={arXiv preprint arXiv:2308.02490},
  year={2023}
}

@article{lu2022learn,
  title={Learn to explain: Multimodal reasoning via thought chains for science question answering},
  author={Lu, Pan and Mishra, Swaroop and Xia, Tanglin and Qiu, Liang and Chang, Kai-Wei and Zhu, Song-Chun and Tafjord, Oyvind and Clark, Peter and Kalyan, Ashwin},
  journal={Advances in Neural Information Processing Systems},
  volume={35},
  pages={2507--2521},
  year={2022}
}

@article{liu2024ocrbench,
  title={Ocrbench: on the hidden mystery of ocr in large multimodal models},
  author={Liu, Yuliang and Li, Zhang and Huang, Mingxin and Yang, Biao and Yu, Wenwen and Li, Chunyuan and Yin, Xu-Cheng and Liu, Cheng-Lin and Jin, Lianwen and Bai, Xiang},
  journal={Science China Information Sciences},
  volume={67},
  number={12},
  pages={220102},
  year={2024},
  publisher={Springer}
}

@inproceedings{mathew2021docvqa,
  title={Docvqa: A dataset for vqa on document images},
  author={Mathew, Minesh and Karatzas, Dimosthenis and Jawahar, CV},
  booktitle={Proceedings of the IEEE/CVF winter conference on applications of computer vision},
  pages={2200--2209},
  year={2021}
}

@article{masry2022chartqa,
  title={Chartqa: A benchmark for question answering about charts with visual and logical reasoning},
  author={Masry, Ahmed and Long, Do Xuan and Tan, Jia Qing and Joty, Shafiq and Hoque, Enamul},
  journal={arXiv preprint arXiv:2203.10244},
  year={2022}
}

@inproceedings{mathew2022infographicvqa,
  title={Infographicvqa},
  author={Mathew, Minesh and Bagal, Viraj and Tito, Rub{\`e}n and Karatzas, Dimosthenis and Valveny, Ernest and Jawahar, CV},
  booktitle={Proceedings of the IEEE/CVF Winter Conference on Applications of Computer Vision},
  pages={1697--1706},
  year={2022}
}

@inproceedings{singh2019towards,
  title={Towards vqa models that can read},
  author={Singh, Amanpreet and Natarajan, Vivek and Shah, Meet and Jiang, Yu and Chen, Xinlei and Batra, Dhruv and Parikh, Devi and Rohrbach, Marcus},
  booktitle={Proceedings of the IEEE/CVF conference on computer vision and pattern recognition},
  pages={8317--8326},
  year={2019}
}

@inproceedings{yu2016modeling,
  title={Modeling context in referring expressions},
  author={Yu, Licheng and Poirson, Patrick and Yang, Shan and Berg, Alexander C and Berg, Tamara L},
  booktitle={European conference on computer vision},
  pages={69--85},
  year={2016},
  organization={Springer}
}

@inproceedings{fu2024blink,
  title={Blink: Multimodal large language models can see but not perceive},
  author={Fu, Xingyu and Hu, Yushi and Li, Bangzheng and Feng, Yu and Wang, Haoyu and Lin, Xudong and Roth, Dan and Smith, Noah A and Ma, Wei-Chiu and Krishna, Ranjay},
  booktitle={European Conference on Computer Vision},
  pages={148--166},
  year={2024},
  organization={Springer}
}

@inproceedings{kembhavi2016diagram,
title={A diagram is worth a dozen images},
author={Kembhavi, Aniruddha and Salvato, Mike and Kolve, Eric and Seo, Minjoon and Hajishirzi, Hannaneh and Farhadi, Ali},
booktitle={European conference on computer vision},
pages={235--251},
year={2016},
organization={Springer}
}

@article{beyer2024paligemma,
      title={{PaliGemma: A versatile 3B VLM for transfer}},
      author={Lucas Beyer and Andreas Steiner and André Susano Pinto and Alexander Kolesnikov and Xiao Wang and Daniel Salz and Maxim Neumann and Ibrahim Alabdulmohsin and Michael Tschannen and Emanuele Bugliarello and Thomas Unterthiner and Daniel Keysers and Skanda Koppula and Fangyu Liu and Adam Grycner and Alexey Gritsenko and Neil Houlsby and Manoj Kumar and Keran Rong and Julian Eisenschlos and Rishabh Kabra and Matthias Bauer and Matko Bošnjak and Xi Chen and Matthias Minderer and Paul Voigtlaender and Ioana Bica and Ivana Balazevic and Joan Puigcerver and Pinelopi Papalampidi and Olivier Henaff and Xi Xiong and Radu Soricut and Jeremiah Harmsen and Xiaohua Zhai},
      year={2024},
      journal={arXiv preprint arXiv:2407.07726}
}

@misc{cheng2025pointarenaprobingmultimodalgrounding,
      title={PointArena: Probing Multimodal Grounding Through Language-Guided Pointing}, 
      author={Long Cheng and Jiafei Duan and Yi Ru Wang and Haoquan Fang and Boyang Li and Yushan Huang and Elvis Wang and Ainaz Eftekhar and Jason Lee and Wentao Yuan and Rose Hendrix and Noah A. Smith and Fei Xia and Dieter Fox and Ranjay Krishna},
      year={2025},
      eprint={2505.09990},
      archivePrefix={arXiv},
      primaryClass={cs.CV},
      url={https://arxiv.org/abs/2505.09990}, 
}

@misc{wiedmann2025finevisionopendataneed,
      title={FineVision: Open Data Is All You Need}, 
      author={Luis Wiedmann and Orr Zohar and Amir Mahla and Xiaohan Wang and Rui Li and Thibaud Frere and Leandro von Werra and Aritra Roy Gosthipaty and Andrés Marafioti},
      year={2025},
      eprint={2510.17269},
      archivePrefix={arXiv},
      primaryClass={cs.CV},
      url={https://arxiv.org/abs/2510.17269}, 
}

@misc{zhang2025beehighqualitycorpusfullstack,
      title={Bee: A High-Quality Corpus and Full-Stack Suite to Unlock Advanced Fully Open MLLMs}, 
      author={Yi Zhang and Bolin Ni and Xin-Sheng Chen and Heng-Rui Zhang and Yongming Rao and Houwen Peng and Qinglin Lu and Han Hu and Meng-Hao Guo and Shi-Min Hu},
      year={2025},
      eprint={2510.13795},
      archivePrefix={arXiv},
      primaryClass={cs.CV},
      url={https://arxiv.org/abs/2510.13795}, 
}

@article{su2024roformer,
  title={Roformer: Enhanced transformer with rotary position embedding},
  author={Su, Jianlin and Ahmed, Murtadha and Lu, Yu and Pan, Shengfeng and Bo, Wen and Liu, Yunfeng},
  journal={Neurocomputing},
  volume={568},
  pages={127063},
  year={2024},
  publisher={Elsevier}
}

@inproceedings{he2022masked,
  title={Masked Autoencoders Are Scalable Vision Learners},
  author={He, Kaiming and Chen, Xinlei and Xie, Saining and Li, Yanghao and Doll{\'a}r, Piotr and Girshick, Ross},
  booktitle={Proceedings of the IEEE/CVF Conference on Computer Vision and Pattern Recognition},
  pages={16000--16009},
  year={2022}
}

@article{tian2024visual,
  title={Visual autoregressive modeling: Scalable image generation via next-scale prediction},
  author={Tian, Keyu and Jiang, Yi and Yuan, Zehuan and Peng, Bingyue and Wang, Liwei},
  journal={Advances in neural information processing systems},
  volume={37},
  pages={84839--84865},
  year={2024}
}

@inproceedings{fang2023eva,
  title={Eva: Exploring the limits of masked visual representation learning at scale},
  author={Fang, Yuxin and Wang, Wen and Xie, Binhui and Sun, Quan and Wu, Ledell and Wang, Xinggang and Huang, Tiejun and Wang, Xinlong and Cao, Yue},
  booktitle={Proceedings of the IEEE/CVF conference on computer vision and pattern recognition},
  pages={19358--19369},
  year={2023}
}

@inproceedings{wei2022mvp,
  title={Mvp: Multimodality-guided visual pre-training},
  author={Wei, Longhui and Xie, Lingxi and Zhou, Wengang and Li, Houqiang and Tian, Qi},
  booktitle={European conference on computer vision},
  pages={337--353},
  year={2022},
  organization={Springer}
}

@article{simeoni2025dinov3,
  title={Dinov3},
  author={Sim{\'e}oni, Oriane and Vo, Huy V and Seitzer, Maximilian and Baldassarre, Federico and Oquab, Maxime and Jose, Cijo and Khalidov, Vasil and Szafraniec, Marc and Yi, Seungeun and Ramamonjisoa, Micha{\"e}l and others},
  journal={arXiv preprint arXiv:2508.10104},
  year={2025}
}

@misc{li2025eagle2buildingposttraining,
      title={Eagle 2: Building Post-Training Data Strategies from Scratch for Frontier Vision-Language Models}, 
      author={Zhiqi Li and Guo Chen and Shilong Liu and Shihao Wang and Vibashan VS and Yishen Ji and Shiyi Lan and Hao Zhang and Yilin Zhao and Subhashree Radhakrishnan and Nadine Chang and Karan Sapra and Amala Sanjay Deshmukh and Tuomas Rintamaki and Matthieu Le and Ilia Karmanov and Lukas Voegtle and Philipp Fischer and De-An Huang and Timo Roman and Tong Lu and Jose M. Alvarez and Bryan Catanzaro and Jan Kautz and Andrew Tao and Guilin Liu and Zhiding Yu},
      year={2025},
      eprint={2501.14818},
      archivePrefix={arXiv},
      primaryClass={cs.CV},
      url={https://arxiv.org/abs/2501.14818}, 
}

@misc{gu2025infinitymmscalingmultimodalperformance,
      title={Infinity-MM: Scaling Multimodal Performance with Large-Scale and High-Quality Instruction Data}, 
      author={Shuhao Gu and Jialing Zhang and Siyuan Zhou and Kevin Yu and Zhaohu Xing and Liangdong Wang and Zhou Cao and Jintao Jia and Zhuoyi Zhang and Yixuan Wang and Zhenchong Hu and Bo-Wen Zhang and Jijie Li and Dong Liang and Yingli Zhao and Songjing Wang and Yulong Ao and Yiming Ju and Huanhuan Ma and Xiaotong Li and Haiwen Diao and Yufeng Cui and Xinlong Wang and Yaoqi Liu and Fangxiang Feng and Guang Liu},
      year={2025},
      eprint={2410.18558},
      archivePrefix={arXiv},
      primaryClass={cs.CL},
      url={https://arxiv.org/abs/2410.18558}, 
}

@misc{li2024llavaonevisioneasyvisualtask,
      title={LLaVA-OneVision: Easy Visual Task Transfer}, 
      author={Bo Li and Yuanhan Zhang and Dong Guo and Renrui Zhang and Feng Li and Hao Zhang and Kaichen Zhang and Peiyuan Zhang and Yanwei Li and Ziwei Liu and Chunyuan Li},
      year={2024},
      eprint={2408.03326},
      archivePrefix={arXiv},
      primaryClass={cs.CV},
      url={https://arxiv.org/abs/2408.03326}, 
}

@inproceedings{cc3m,
    title = "Conceptual Captions: A Cleaned, Hypernymed, Image Alt-text Dataset For Automatic Image Captioning",
    author = "Sharma, Piyush  and
      Ding, Nan  and
      Goodman, Sebastian  and
      Soricut, Radu",
    editor = "Gurevych, Iryna  and
      Miyao, Yusuke",
    booktitle = "Proceedings of the 56th Annual Meeting of the Association for Computational Linguistics (Volume 1: Long Papers)",
    month = jul,
    year = "2018",
    address = "Melbourne, Australia",
    publisher = "Association for Computational Linguistics",
    url = "https://aclanthology.org/P18-1238/",
    doi = "10.18653/v1/P18-1238",
    pages = "2556--2565"
}

@inproceedings{changpinyo2021cc12m,
  title = {{Conceptual 12M}: Pushing Web-Scale Image-Text Pre-Training To Recognize Long-Tail Visual Concepts},
  author = {Changpinyo, Soravit and Sharma, Piyush and Ding, Nan and Soricut, Radu},
  booktitle = {CVPR},
  year = {2021},
}

@InProceedings{coco,
author="Lin, Tsung-Yi
and Maire, Michael
and Belongie, Serge
and Hays, James
and Perona, Pietro
and Ramanan, Deva
and Doll{\'a}r, Piotr
and Zitnick, C. Lawrence",
editor="Fleet, David
and Pajdla, Tomas
and Schiele, Bernt
and Tuytelaars, Tinne",
title="Microsoft COCO: Common Objects in Context",
booktitle="Computer Vision -- ECCV 2014",
year="2014",
publisher="Springer International Publishing",
address="Cham",
pages="740--755"
}

@misc{kakaobrain2022coyo-700m,
  title         = {COYO-700M: Image-Text Pair Dataset},
  author        = {Byeon, Minwoo and Park, Beomhee and Kim, Haecheon and Lee, Sungjun and Baek, Woonhyuk and Kim, Saehoon},
  year          = {2022},
  howpublished  = {\url{https://github.com/kakaobrain/coyo-dataset}},
}

@inproceedings{
gadre2023datacomp,
title={DataComp: In search of the next generation of multimodal datasets},
author={Samir Yitzhak Gadre and Gabriel Ilharco and Alex Fang and Jonathan Hayase and Georgios Smyrnis and Thao Nguyen and Ryan Marten and Mitchell Wortsman and Dhruba Ghosh and Jieyu Zhang and Eyal Orgad and Rahim Entezari and Giannis Daras and Sarah M Pratt and Vivek Ramanujan and Yonatan Bitton and Kalyani Marathe and Stephen Mussmann and Richard Vencu and Mehdi Cherti and Ranjay Krishna and Pang Wei Koh and Olga Saukh and Alexander Ratner and Shuran Song and Hannaneh Hajishirzi and Ali Farhadi and Romain Beaumont and Sewoong Oh and Alex Dimakis and Jenia Jitsev and Yair Carmon and Vaishaal Shankar and Ludwig Schmidt},
booktitle={Thirty-seventh Conference on Neural Information Processing Systems Datasets and Benchmarks Track},
year={2023},
url={https://openreview.net/forum?id=dVaWCDMBof}
}

@inproceedings{wukong,
 author = {Gu, Jiaxi and Meng, Xiaojun and Lu, Guansong and Hou, Lu and Minzhe, Niu and Liang, Xiaodan and Yao, Lewei and Huang, Runhui and Zhang, Wei and Jiang, Xin and XU, Chunjing and Xu, Hang},
 booktitle = {Advances in Neural Information Processing Systems},
 editor = {S. Koyejo and S. Mohamed and A. Agarwal and D. Belgrave and K. Cho and A. Oh},
 pages = {26418--26431},
 publisher = {Curran Associates, Inc.},
 title = {Wukong: A 100 Million Large-scale Chinese Cross-modal Pre-training Benchmark},
 url = {https://proceedings.neurips.cc/paper_files/paper/2022/file/a90b9a09a6ee43d6631cf42e225d73b4-Paper-Datasets_and_Benchmarks.pdf},
 volume = {35},
 year = {2022}
}

@inproceedings{sbu,
 author = {Ordonez, Vicente and Kulkarni, Girish and Berg, Tamara},
 booktitle = {Advances in Neural Information Processing Systems},
 editor = {J. Shawe-Taylor and R. Zemel and P. Bartlett and F. Pereira and K.Q. Weinberger},
 pages = {},
 publisher = {Curran Associates, Inc.},
 title = {Im2Text: Describing Images Using 1 Million Captioned Photographs},
 url = {https://proceedings.neurips.cc/paper/2011/file/5dd9db5e033da9c6fb5ba83c7a7ebea9-Paper.pdf},
 volume = {24},
 year = {2011}
}

@article{flickr,
  title={From image descriptions to visual denotations: New similarity metrics for semantic inference over event descriptions},
  author={Young, Peter and Lai, Alice and Hodosh, Micah and Hockenmaier, Julia},
  journal={Transactions of the Association for Computational Linguistics},
  volume={2},
  pages={67--78},
  year={2014},
  publisher={MIT Press}
}

@inproceedings{zhang2024long,
  title={Long-clip: Unlocking the long-text capability of clip},
  author={Zhang, Beichen and Zhang, Pan and Dong, Xiaoyi and Zang, Yuhang and Wang, Jiaqi},
  booktitle={European conference on computer vision},
  pages={310--325},
  year={2024},
  organization={Springer}
}

@inproceedings{wang2025fix,
  title={Fix-clip: Dual-branch hierarchical contrastive learning via synthetic captions for better understanding of long text},
  author={Wang, Bingchao and Ning, Zhiwei and Ding, Jianyu and Gao, Xuanang and Li, Yin and Jiang, Dongsheng and Yang, Jie and Liu, Wei},
  booktitle={Proceedings of the IEEE/CVF International Conference on Computer Vision},
  pages={20694--20704},
  year={2025}
}

@article{wang2025internvl3,
  title={Internvl3. 5: Advancing open-source multimodal models in versatility, reasoning, and efficiency},
  author={Wang, Weiyun and Gao, Zhangwei and Gu, Lixin and Pu, Hengjun and Cui, Long and Wei, Xingguang and Liu, Zhaoyang and Jing, Linglin and Ye, Shenglong and Shao, Jie and others},
  journal={arXiv preprint arXiv:2508.18265},
  year={2025}
}

@article{gu2024infinity,
  title={Infinity-mm: Scaling multimodal performance with large-scale and high-quality instruction data},
  author={Gu, Shuhao and Zhang, Jialing and Zhou, Siyuan and Yu, Kevin and Xing, Zhaohu and Wang, Liangdong and Cao, Zhou and Jia, Jintao and Zhang, Zhuoyi and Wang, Yixuan and others},
  journal={arXiv preprint arXiv:2410.18558},
  year={2024}
}

@article{Qwen3-VL,
      title={Qwen3-VL Technical Report}, 
      author={Shuai Bai and Yuxuan Cai and Ruizhe Chen and Keqin Chen and Xionghui Chen and Zesen Cheng and Lianghao Deng and Wei Ding and Chang Gao and Chunjiang Ge and Wenbin Ge and Zhifang Guo and Qidong Huang and Jie Huang and Fei Huang and Binyuan Hui and Shutong Jiang and Zhaohai Li and Mingsheng Li and Mei Li and Kaixin Li and Zicheng Lin and Junyang Lin and Xuejing Liu and Jiawei Liu and Chenglong Liu and Yang Liu and Dayiheng Liu and Shixuan Liu and Dunjie Lu and Ruilin Luo and Chenxu Lv and Rui Men and Lingchen Meng and Xuancheng Ren and Xingzhang Ren and Sibo Song and Yuchong Sun and Jun Tang and Jianhong Tu and Jianqiang Wan and Peng Wang and Pengfei Wang and Qiuyue Wang and Yuxuan Wang and Tianbao Xie and Yiheng Xu and Haiyang Xu and Jin Xu and Zhibo Yang and Mingkun Yang and Jianxin Yang and An Yang and Bowen Yu and Fei Zhang and Hang Zhang and Xi Zhang and Bo Zheng and Humen Zhong and Jingren Zhou and Fan Zhou and Jing Zhou and Yuanzhi Zhu and Ke Zhu},
	  journal={arXiv preprint arXiv:2511.21631},
      year={2025}
}

@inproceedings{deng2009imagenet,
  title={Imagenet: A large-scale hierarchical image database},
  author={Deng, Jia and Dong, Wei and Socher, Richard and Li, Li-Jia and Li, Kai and Fei-Fei, Li},
  booktitle={2009 IEEE conference on computer vision and pattern recognition},
  pages={248--255},
  year={2009},
  organization={Ieee}
}

@inproceedings{recht2019imagenet,
  title={Do imagenet classifiers generalize to imagenet?},
  author={Recht, Benjamin and Roelofs, Rebecca and Schmidt, Ludwig and Shankar, Vaishaal},
  booktitle={International conference on machine learning},
  pages={5389--5400},
  year={2019},
  organization={PMLR}
}

@article{beyer2020we,
  title={Are we done with imagenet?},
  author={Beyer, Lucas and H{\'e}naff, Olivier J and Kolesnikov, Alexander and Zhai, Xiaohua and Oord, A{\"a}ron van den},
  journal={arXiv preprint arXiv:2006.07159},
  year={2020}
}

@article{chen2015microsoft,
  title={Microsoft coco captions: Data collection and evaluation server},
  author={Chen, Xinlei and Fang, Hao and Lin, Tsung-Yi and Vedantam, Ramakrishna and Gupta, Saurabh and Doll{\'a}r, Piotr and Zitnick, C Lawrence},
  journal={arXiv preprint arXiv:1504.00325},
  year={2015}
}

@inproceedings{plummer2015flickr30k,
  title={Flickr30k entities: Collecting region-to-phrase correspondences for richer image-to-sentence models},
  author={Plummer, Bryan A and Wang, Liwei and Cervantes, Chris M and Caicedo, Juan C and Hockenmaier, Julia and Lazebnik, Svetlana},
  booktitle={Proceedings of the IEEE international conference on computer vision},
  pages={2641--2649},
  year={2015}
}

@inproceedings{urbanek2024picture,
  title={A picture is worth more than 77 text tokens: Evaluating clip-style models on dense captions},
  author={Urbanek, Jack and Bordes, Florian and Astolfi, Pietro and Williamson, Mary and Sharma, Vasu and Romero-Soriano, Adriana},
  booktitle={Proceedings of the IEEE/CVF Conference on Computer Vision and Pattern Recognition},
  pages={26700--26709},
  year={2024}
}

@inproceedings{garg2024imageinwords,
  title={Imageinwords: Unlocking hyper-detailed image descriptions},
  author={Garg, Roopal and Burns, Andrea and Karagol-Ayan, Burcu and Bitton, Yonatan and Montgomery, Ceslee and Onoe, Yasumasa and Bunner, Andrew and Krishna, Ranjay and Baldridge, Jason Michael and Soricut, Radu},
  booktitle={Proceedings of the 2024 Conference on Empirical Methods in Natural Language Processing},
  pages={93--127},
  year={2024}
}

@article{liu2024visualwebbench,
  title={Visualwebbench: How far have multimodal llms evolved in web page understanding and grounding?},
  author={Liu, Junpeng and Song, Yifan and Lin, Bill Yuchen and Lam, Wai and Neubig, Graham and Li, Yuanzhi and Yue, Xiang},
  journal={arXiv preprint arXiv:2404.05955},
  year={2024}
}

@inproceedings{ranjan2021learning,
  title={Learning to count everything},
  author={Ranjan, Viresh and Sharma, Udbhav and Nguyen, Thu and Hoai, Minh},
  booktitle={Proceedings of the IEEE/CVF conference on computer vision and pattern recognition},
  pages={3394--3403},
  year={2021}
}

@inproceedings{duan2024vlmevalkit,
  title={Vlmevalkit: An open-source toolkit for evaluating large multi-modality models},
  author={Duan, Haodong and Yang, Junming and Qiao, Yuxuan and Fang, Xinyu and Chen, Lin and Liu, Yuan and Dong, Xiaoyi and Zang, Yuhang and Zhang, Pan and Wang, Jiaqi and others},
  booktitle={Proceedings of the 32nd ACM international conference on multimedia},
  pages={11198--11201},
  year={2024}
}

\clearpage
\appendix
\section*{\centering Appendix}
\section{Data Statistics}
This section details the construction and filtering protocols used to develop the multi-stage training corpus for FineViT, alongside granular distribution analyses and visualizations. FineViT employs a progressive data refinement strategy across its training trajectory. As delineated in Table~\ref{tab:training_data_more}, we curated this extensive dataset by aggregating and processing diverse open-source and internal data sources to ensure high data quality and distributional diversity.

\begin{table}[h!]
    \centering
    \caption{Detailed information of the data at different FineViT training stages.}
    \label{tab:training_data_more}
    \resizebox{\textwidth}{!}{%
    \setlength{\tabcolsep}{6pt} 
    \renewcommand{\arraystretch}{1.4}
    \begin{tabular}{l c c c}
        \toprule
        \textbf{Stage} & \textbf{Rules} & \textbf{Data Source} & \textbf{Data Size} \\
        \midrule
        \makecell{Stage I} & \makecell{\textbf{image}: short $\geqslant$ 224, \\ remove duplicates} & \multirow{2}{*}{\makecell{COCO~\cite{coco}, CC3M~\cite{cc3m}, CC12M~\cite{changpinyo2021cc12m}, \\ YFCC15M~\cite{radford2021learning}, LAION~\cite{schuhmann2021laion}, \\ wukong~\cite{wukong}, datacomp~\cite{gadre2023datacomp}, COYO~\cite{kakaobrain2022coyo-700m}, \\ SBU~\cite{sbu}, Flickr30k~\cite{flickr}}} & \makecell{1.8B images} \\ \cmidrule(r){1-2} \cmidrule(l){4-4}
        \makecell{Stage II} & \makecell{\textbf{caption}: remove \\
repeated or null captions} & & 1.56B image-text pairs\\
        \midrule
        \makecell{Stage III} & \makecell{\textbf{image}: short $\geqslant$ 448, \\ ratio $\in$ [1/3, 3], \\ Blurry, Dark, Saturation check \\ \textbf{region}: bbox area $\geqslant$ $1\%$, \\ confidence scores $\geqslant$ 0.3, \\ NMS with $\text{IoU} = 0.7$, \\ label balance}& \makecell{previous stage data; \\ additional OCR data from: \\ FineVision~\cite{wiedmann2025finevisionopendataneed}, LLaVA-OneVision~\cite{li2024llavaonevisioneasyvisualtask}, \\ Eagle 2~\cite{li2025eagle2buildingposttraining}, Infinity-MM~\cite{gu2025infinitymmscalingmultimodalperformance}, \\ LLaVA-OneVision-1.5-Instruct~\cite{LLaVA-OneVision-1.5} }& \makecell{80M images \\ 454M regions} \\
        \bottomrule
    \end{tabular}}
\end{table}

$\bullet$ \textbf{Stage I (1.8B images)}: We initialised our corpus with $1.8$B images sourced from a comprehensive mixture of datasets, including COCO, CC12M, LAION, and Datacomp. To alleviate the impact of low-quality and redundant samples, we filtered the corpus by a minimum spatial dimension of $224$ pixels and performed p-Hash-based deduplication, ensuring a diverse and high-fidelity training distribution.
  
$\bullet$ \textbf{Stage II (1.56B iamge-text pairs)}: Building upon the initial corpus, we refined image-text alignment through a large-scale synthetic recaptioning pipeline. To mitigate the inductive biases of individual models, the $1.8$B samples were processed via an ensemble of three MLLMs: Qwen2.5-7B \cite{bai2025qwen2}, Intern3-VL-8B \cite{zhu2025internvl3}, and MiniCPM-V-8B \cite{yao2024minicpm}, with images stochastically assigned to each. By filtering out non-informative, null, or repetitive captions, we distilled the dataset into $1.56$B high-quality image-text pairs, thereby alleviating potential noise during the contrastive learning phase.

$\bullet$ \textbf{Stage III (80M images \& 454M regions)}: To facilitate fine-grained visual perception, we prioritize high-resolution samples ($\ge$ $448$ pixels) within a constrained aspect ratio range of $[1/3, 3]$. This corpus underwent a rigorous perceptual quality assessment to exclude degraded samples, including those with significant motion blur, extreme low-light conditions, or oversaturation. For region-level localization, we first extract noun phrases from global recaptions and utilize Grounding-DINO \cite{liu2024grounding} to generate candidate bounding boxes. To ensure localization quality, we prune candidates with confidence scores below $0.3$ or those occupying less than $1\%$ of the total image area. Redundant detections are further consolidated via class-agnostic Non-Maximum Suppression (NMS) with an IoU threshold of $0.7$. To address class imbalance, we implement a stratified frequency-based sampling strategy across 10M-image batches. Categories with fewer than $1,000$ instances are fully retained, while over-represented classes are downsampled proportionally, capped at a maximum of $100,000$ instances per batch. This ensures a more uniform distribution across the taxonomy. To generate context-aware local descriptions, we adopt a dual-stream prompting strategy inspired by \cite{shi2025scaling}. By feeding both the localized crop and the global context into the Qwen3-VL 32B Model, we elicit captions that capture intricate local details while maintaining global semantic coherence.
This pipeline culminated in a curated dataset comprising $80$M images and $454$M dense region-text pairs, providing a robust foundation for high-fidelity spatial reasoning.


\subsection{Data Statistics of Stage II: 1.56B images}
\begin{figure}[t]
  \centering
  \includegraphics[width=0.95\columnwidth]{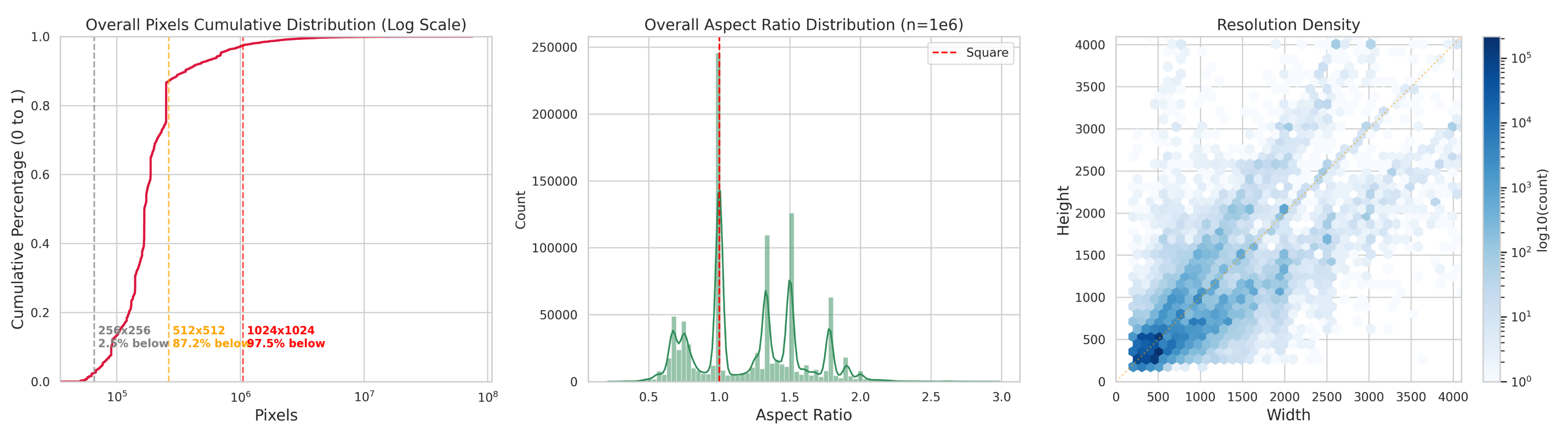}
  \caption{Resolution distribution of $1.56$B images used at stage II.}
  \label{fig:resolution_distribution}
\end{figure}

\begin{wrapfigure}{r}{0.36\textwidth}
  \centering
  \vspace{-8pt}
  \includegraphics[width=0.35\textwidth]{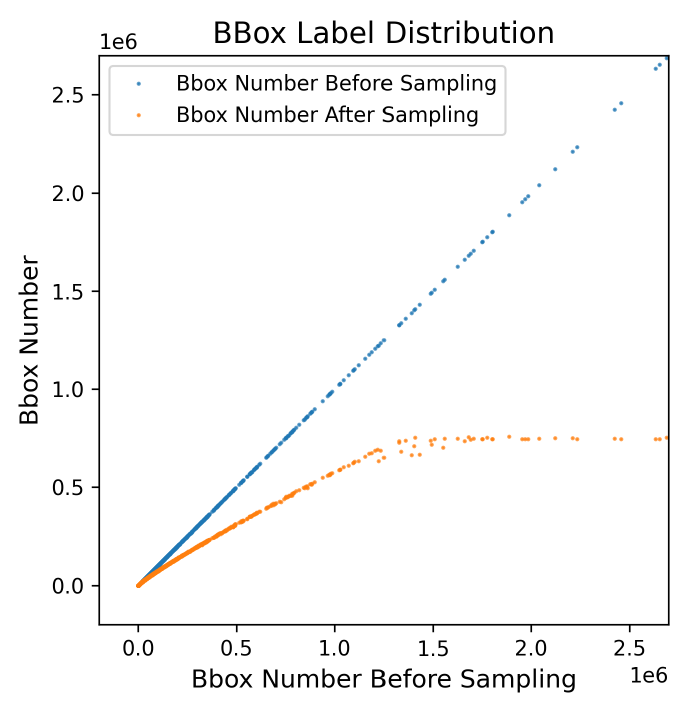}
  \caption{Category distribution.}
  \label{fig:box_label}
\end{wrapfigure}
As shown in Figure~\ref{fig:resolution_distribution}, given the vast scale of our full dataset, we performed a comprehensive statistical analysis on a representative subset of $10^6$ samples to characterize the underlying data distribution. The resolution density plot reveals a diverse spatial range, with image dimensions capped at $4096$ pixels. Analysis of the cumulative pixel distribution indicates that $2.5\%$, $87.2\%$, and $97.5\%$ of the images fall below the thresholds of $256 \times 256$, $512 \times 512$, and $1024 \times 1024$, respectively. This hierarchical distribution ensures a balanced representation of both low-resolution features and high-definition details. Furthermore, the aspect ratio distribution exhibits a prominent peak at $1.0$ (square) while spanning a broad spectrum (capped at $3.0$ for visualization). This geometric diversity enables the model to generalize effectively across the various framing and cropping scenarios typical of real-world applications.

\subsection{Data Statistics of Stage III: FineCap 450M}

This section introduces the class-balanced sampling of region bboxes, as well as the distribution of region bboxes and their corresponding text.


\noindent\textbf{Region-level Category Distribution.}
Figure~\ref{fig:box_label} illustrates the distribution of bounding boxes across categories before and after sampling. Notably, the post-sampling curve exhibits a more balanced profile, significantly mitigating class imbalance. Following this sampling strategy, we obtained $226$M bounding boxes of gereral regions spanning $63,125$ classes. By integrating $142$M general OCR and 86M Document OCR boxes, the final dataset comprises $454$M bounding boxes across $80$M images (consisting of $63$M general and $17$M OCR-specific images). 

\begin{figure}[h!]
  \centering
  \includegraphics[width=0.24\columnwidth]{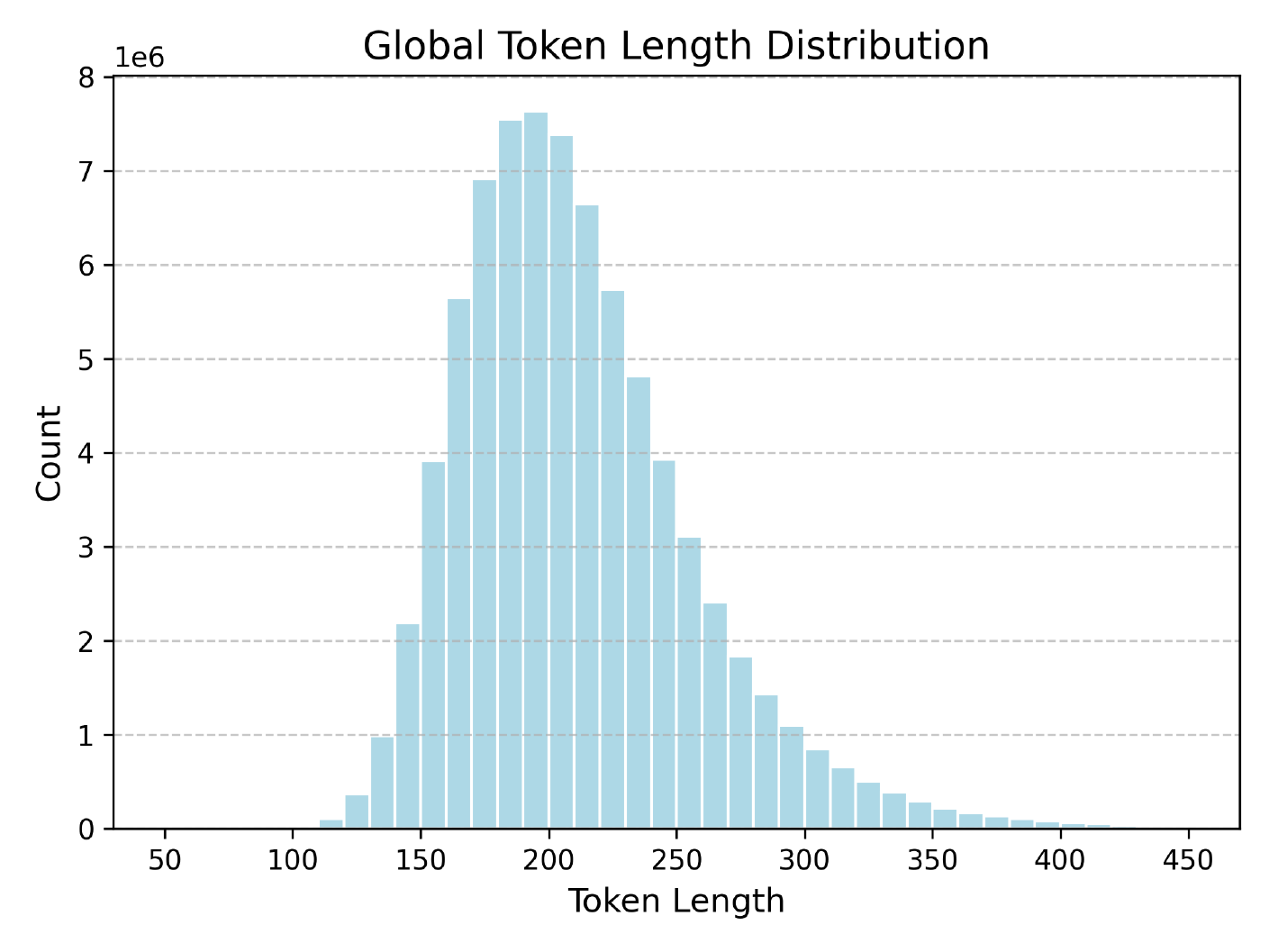}
  \includegraphics[width=0.24\columnwidth]{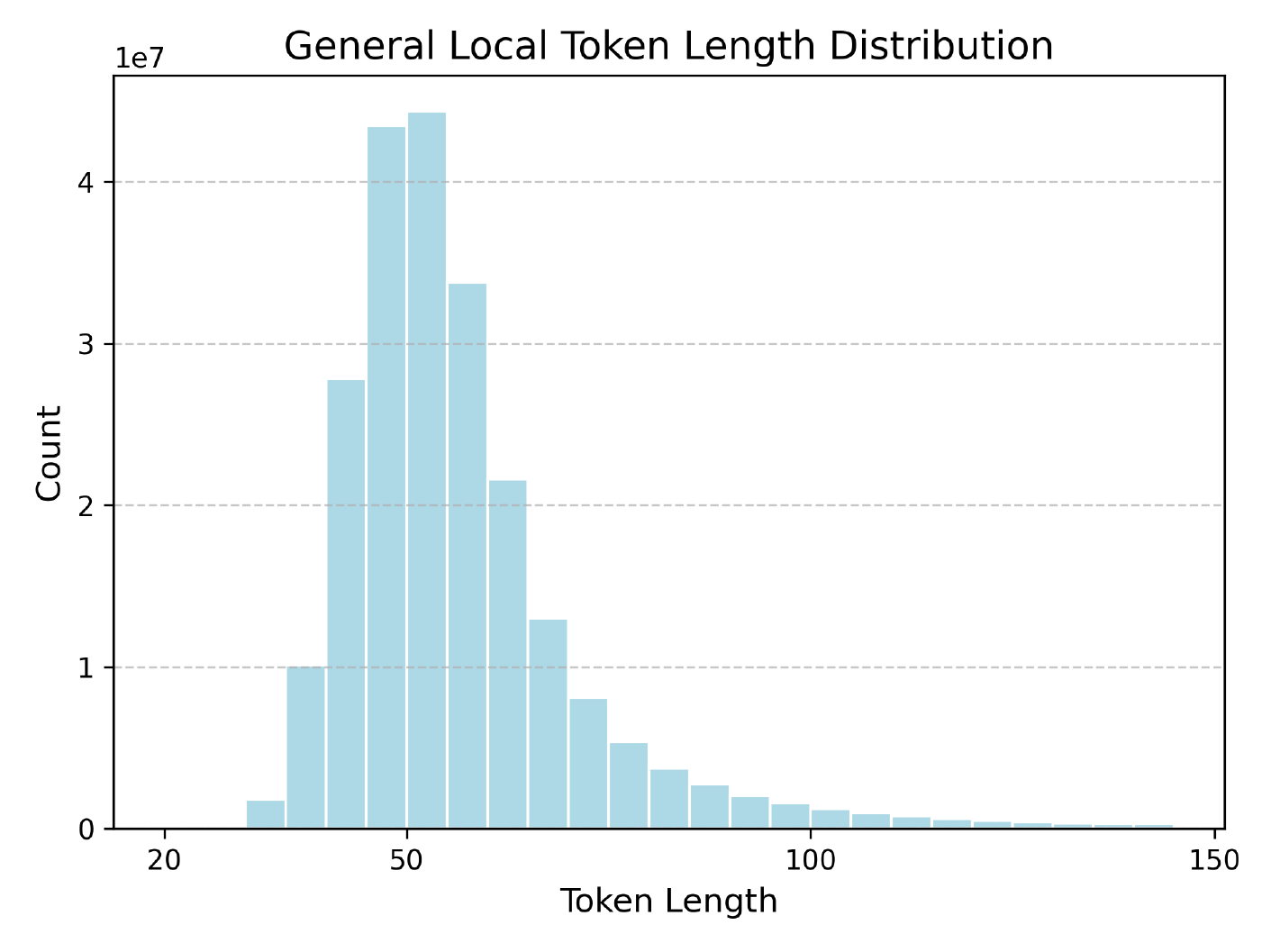}
  \includegraphics[width=0.24\columnwidth]{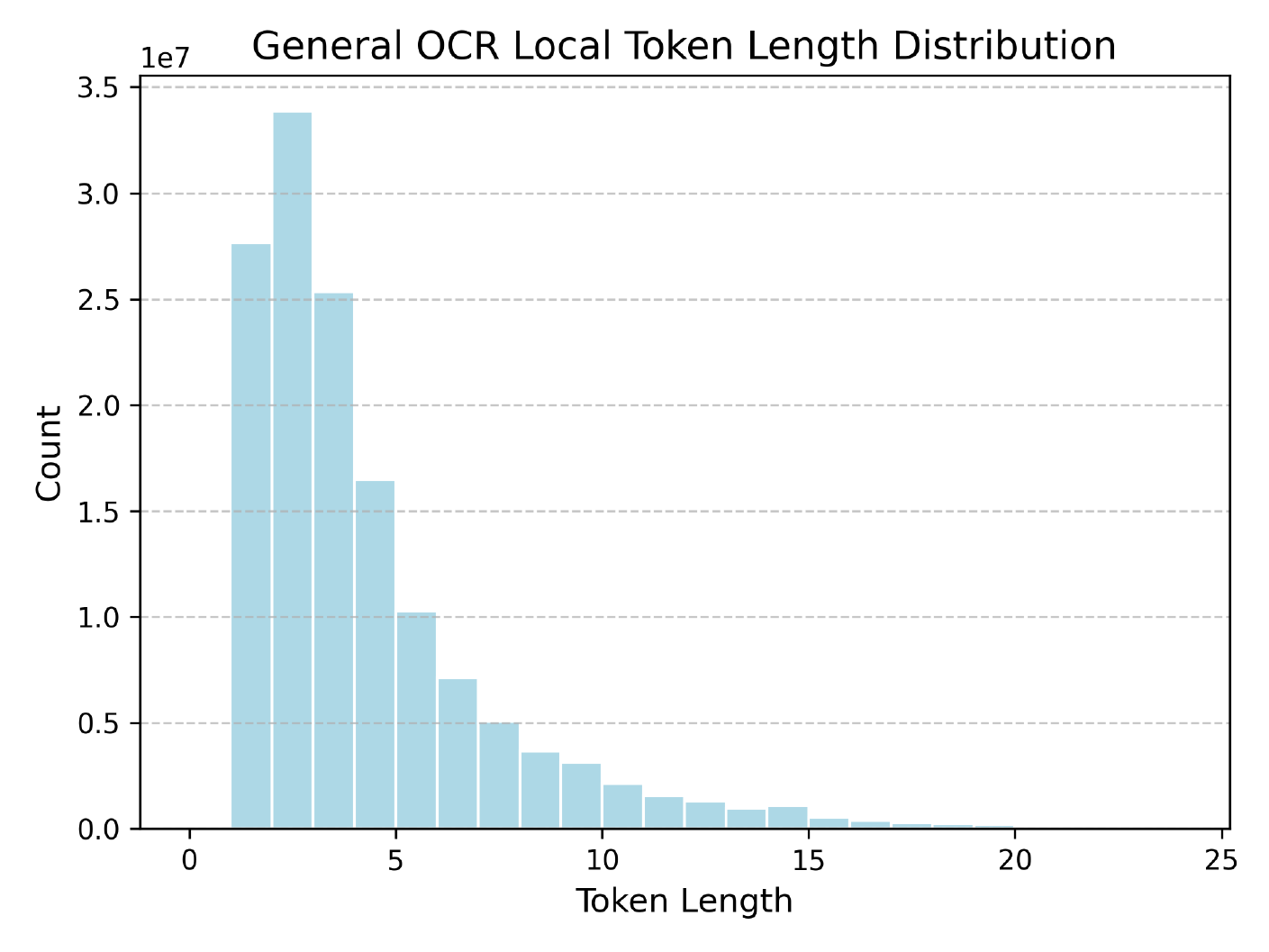}
  \includegraphics[width=0.24\columnwidth]{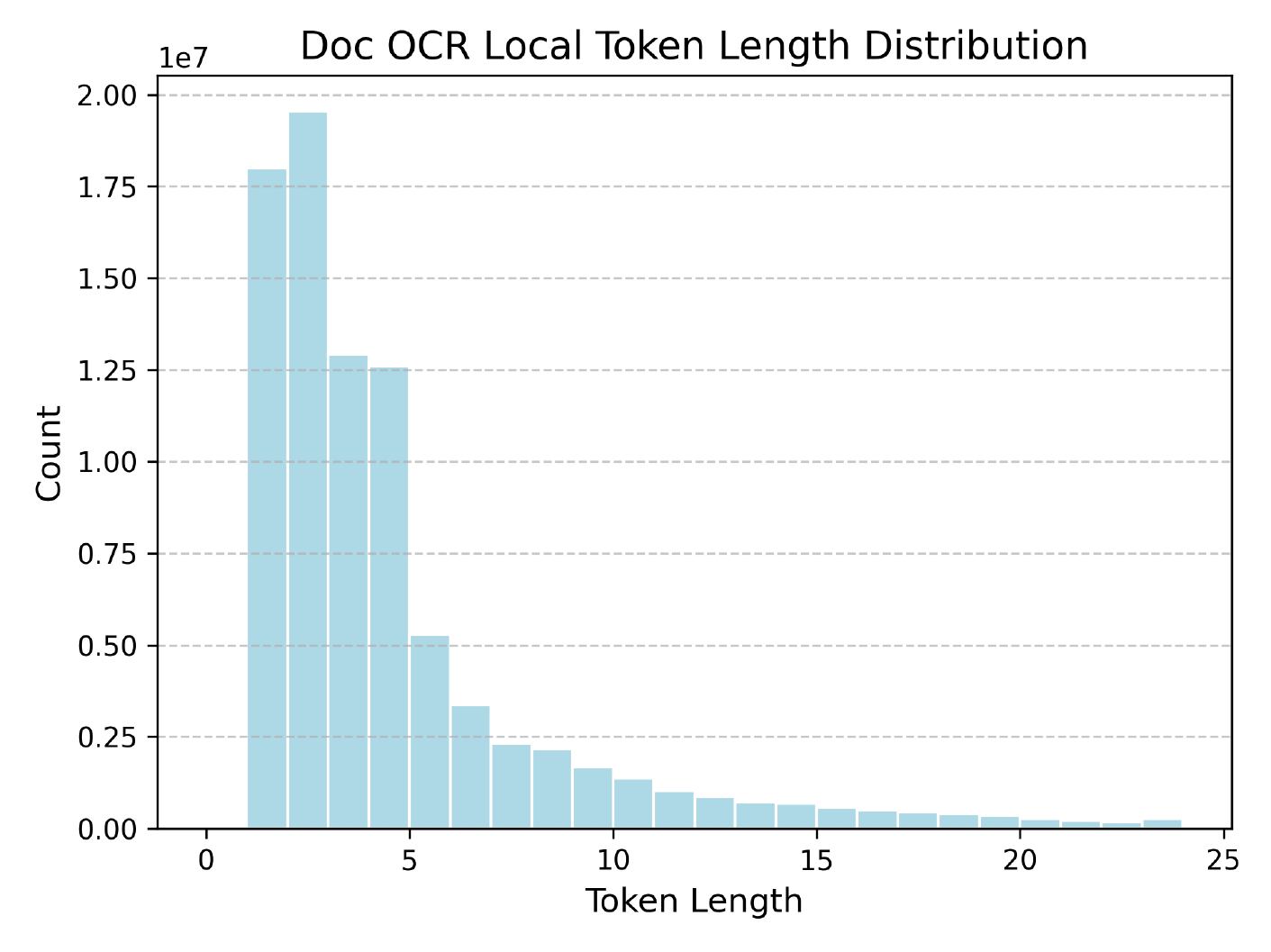}
  \caption{Token Length Distribution of of FineCap-450M.}
  \label{fig:token_length}
\end{figure}

\noindent\textbf{Token Length Distribution.} Figure~\ref{fig:token_length} illustrates the token length distributions for global and local captions, as well as general and document OCR. Specifically, global captions typically span between $150$ and $300$ tokens, whereas local captions cluster near $50$. In contrast, OCR-related tokens are significantly shorter, with the majority falling below $10$. 

\begin{figure}[h!]
  \centering
  \includegraphics[width=0.3\columnwidth]{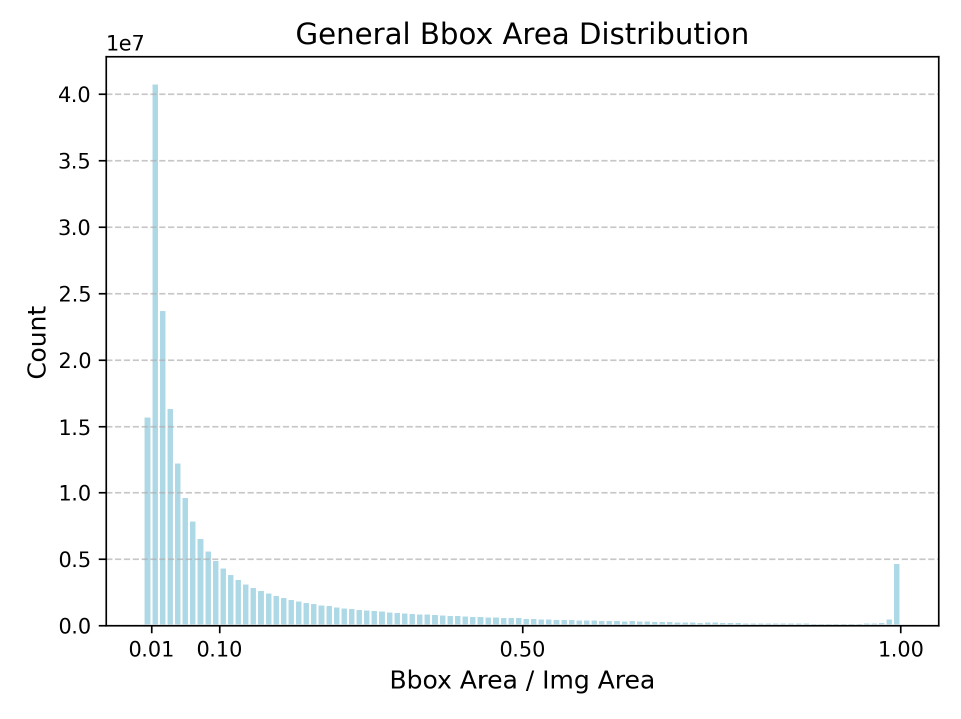}
  \includegraphics[width=0.3\columnwidth]{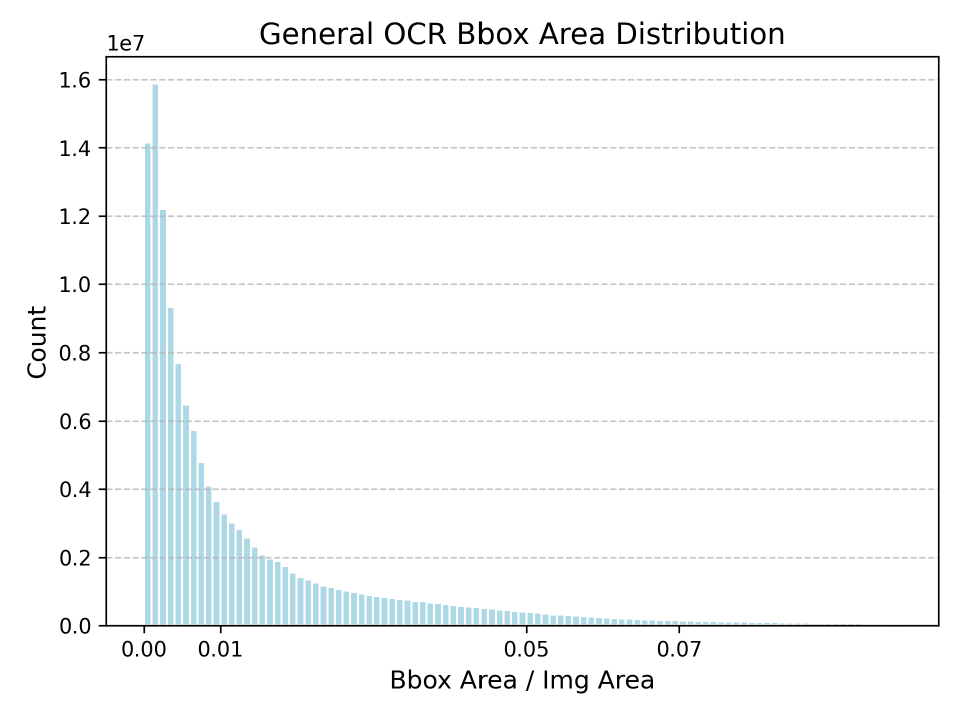}
  \includegraphics[width=0.3\columnwidth]{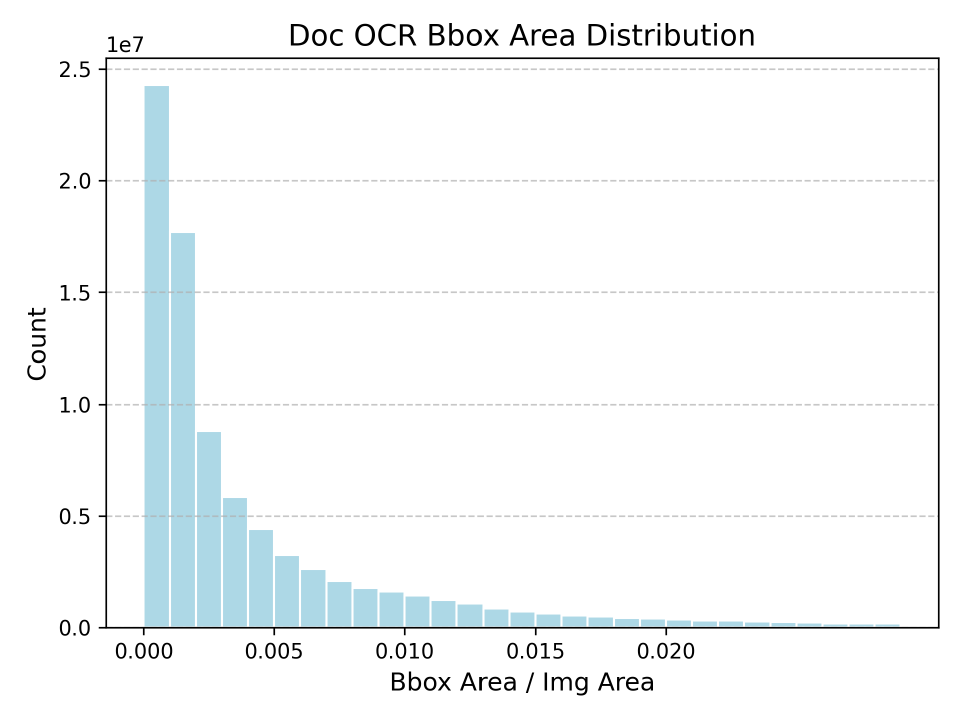}
  \caption{Relative bounding box area distribution of FineCap-450M.}
  \label{fig:box_area}
\end{figure}

\noindent\textbf{Relative Bounding Box Area Distribution.} Figure~\ref{fig:box_area} illustrates the distribution of the area ratios for the bounding boxes of general, general OCR, and document OCR relative to the entire image. The area ratios of general bounding boxes are predominantly concentrated below $10\%$ or near $100\%$. In contrast, owing to the inherent nature of textual content, OCR-derived bounding boxes typically occupy a much smaller proportion of the total image area. 

\begin{figure}[h!]
  \centering
  \includegraphics[width=0.3\columnwidth]{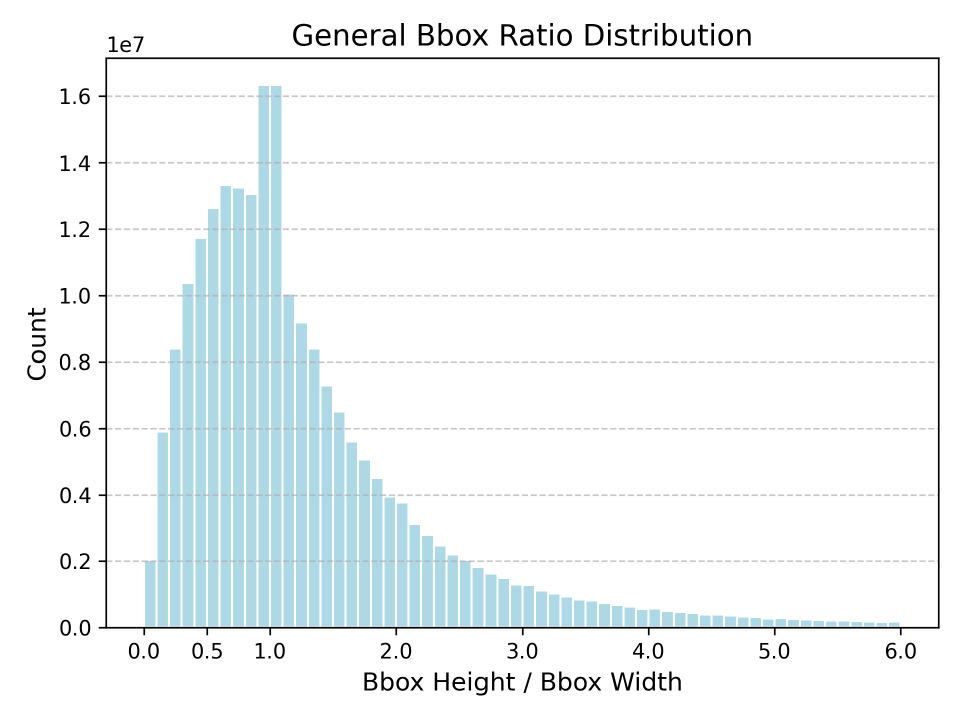}
  \includegraphics[width=0.3\columnwidth]{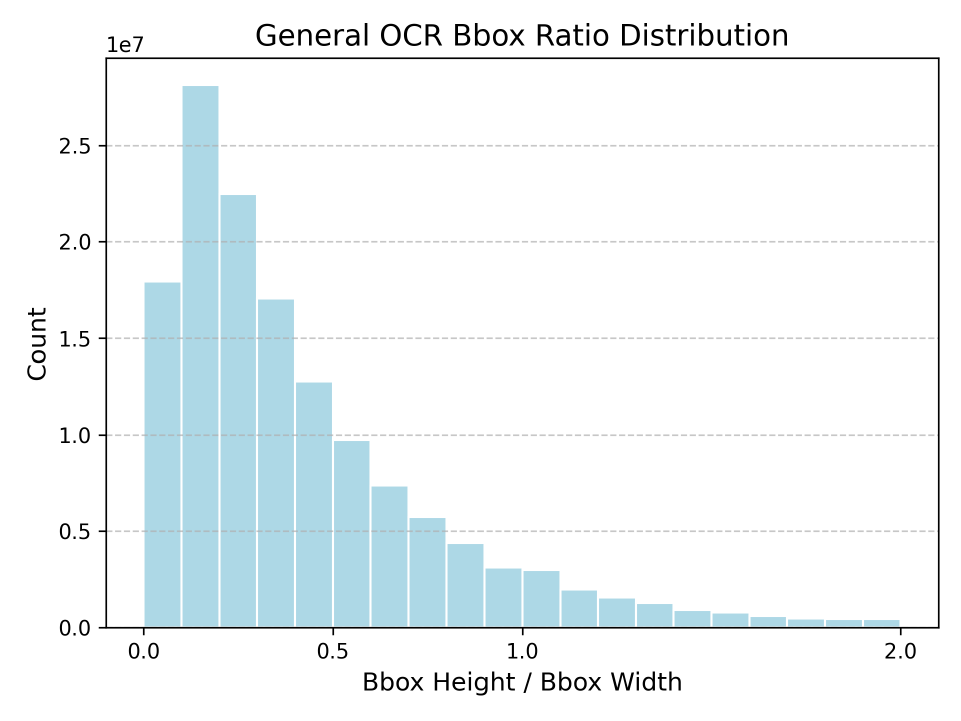}
  \includegraphics[width=0.3\columnwidth]{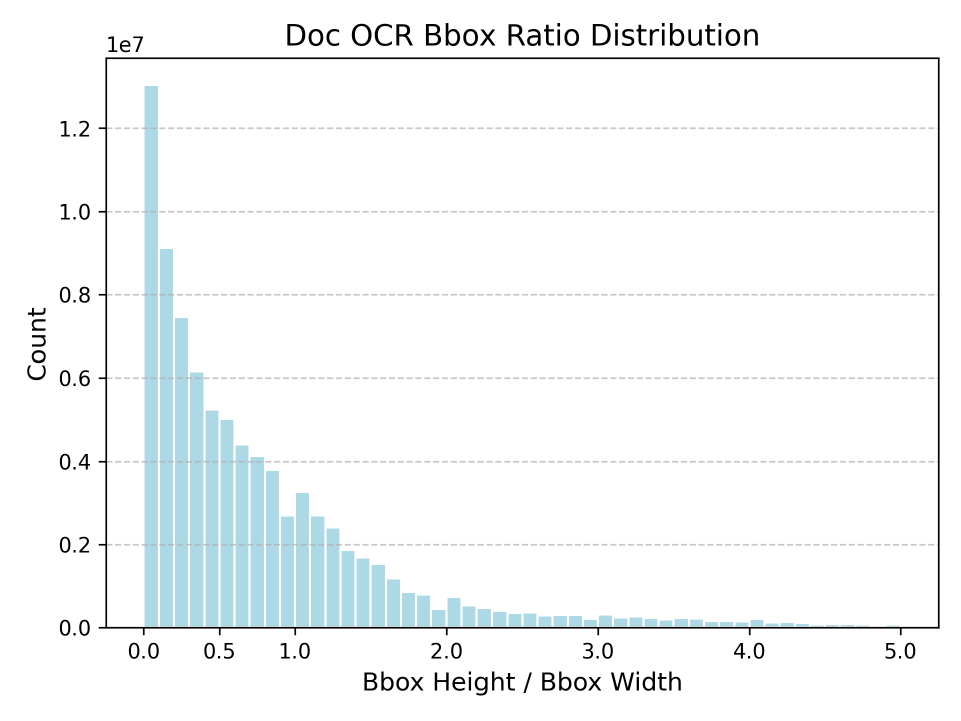}
  \caption{Bounding box aspect ratio distribution of FineCap-450M.}
  \label{fig:box_ratio}
\end{figure}

\noindent\textbf{Bounding Box Aspect Ratio Distribution.} Figure~\ref{fig:box_ratio} illustrates the aspect ratio distributions across different types of bounding box (general, general OCR, and Doc OCR). General bounding boxes exhibit a relatively balanced distribution, with a noticeable concentration around an aspect ratio of 1:1 (equilateral). In contrast, OCR-derived bounding boxes are predominantly elongated, reflecting the inherent linear structure of textual content. 

\subsection{Illustration of FineCap-450M}

\begin{figure}[h!]
  \centering
  \includegraphics[width=0.95\columnwidth]{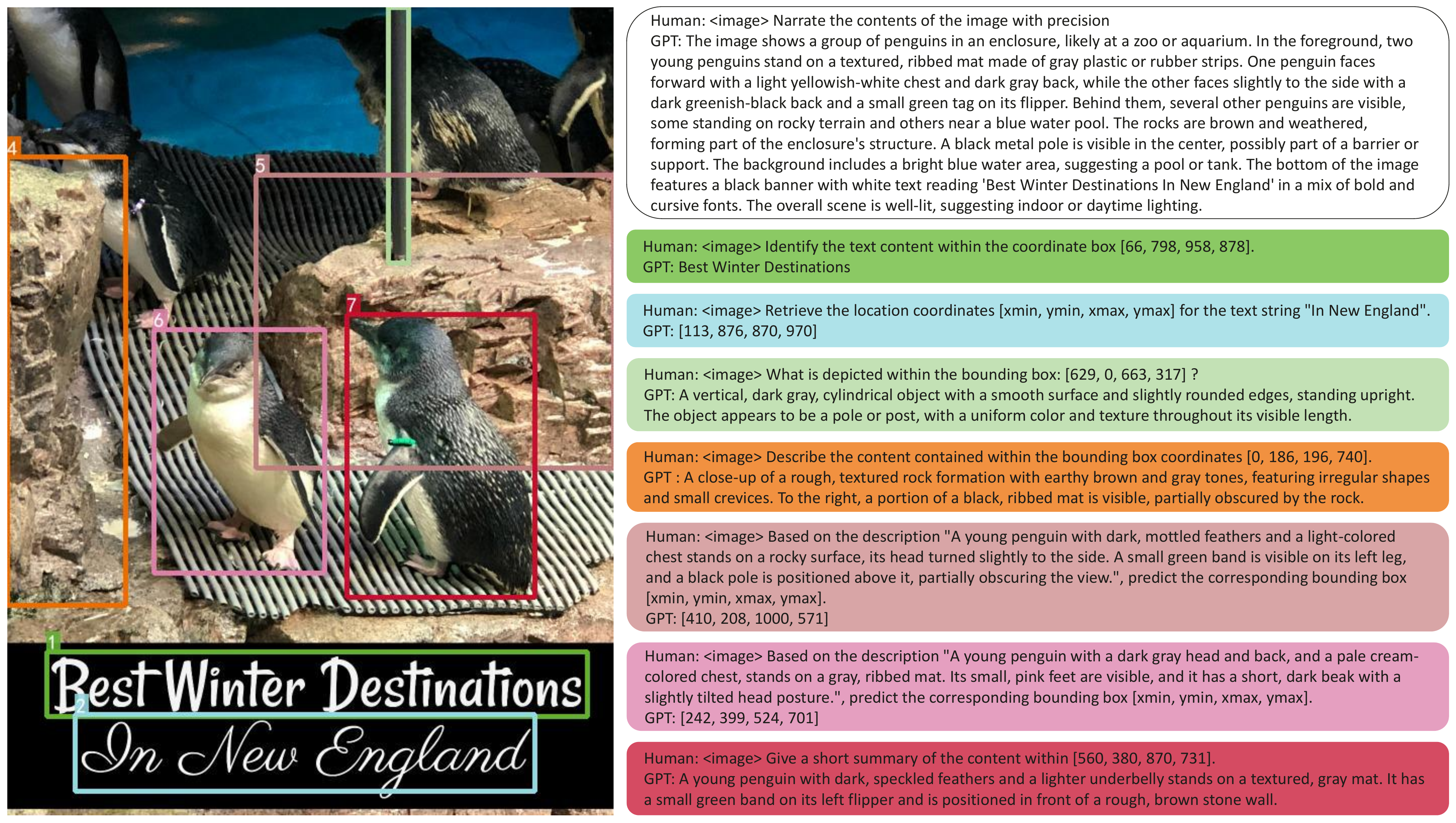}
  \includegraphics[width=0.95\columnwidth]{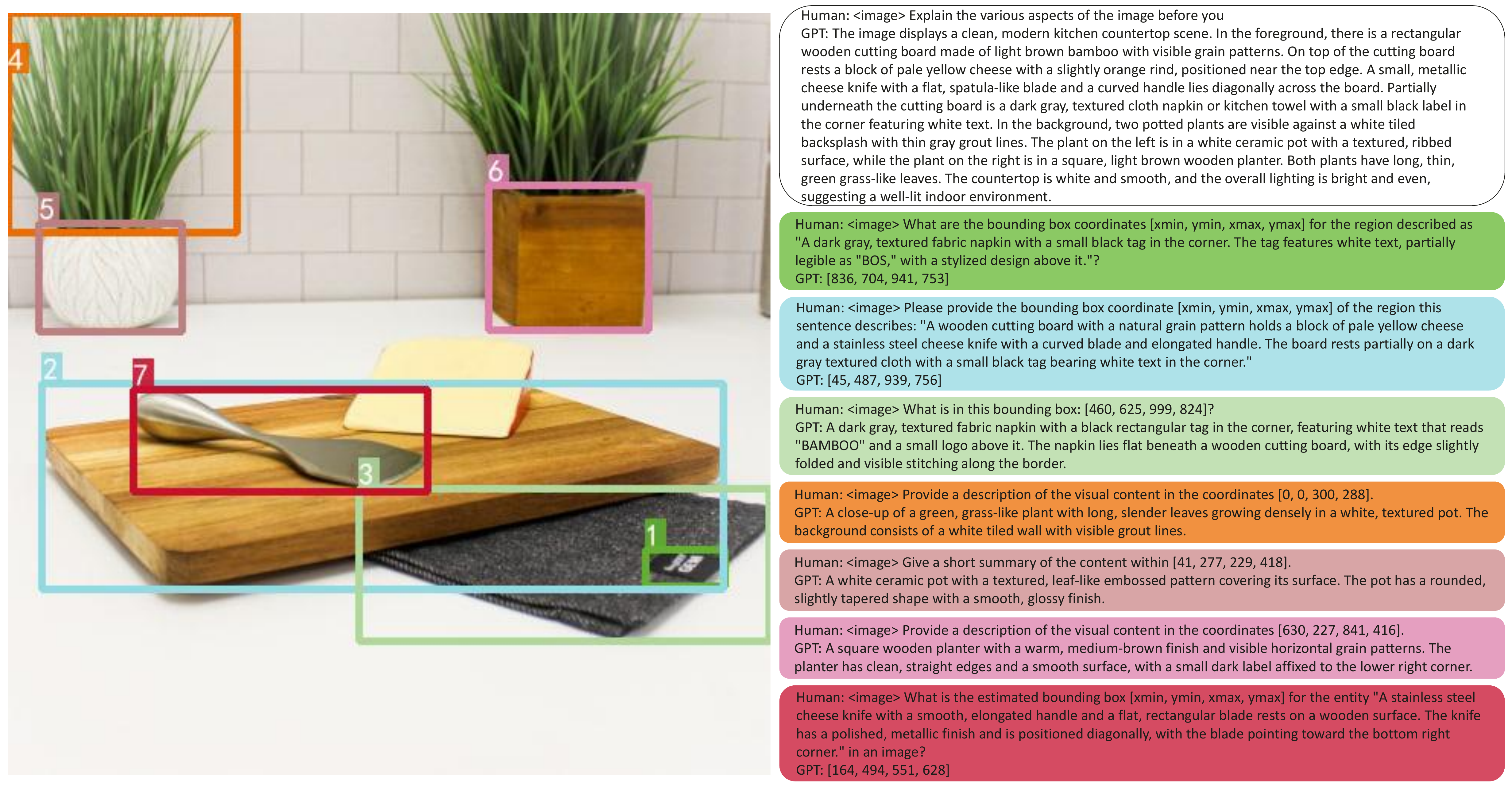}
  \caption{Illustration of FineCap-450M. To prevent an excessive number of regions, we randomly sample a subset for better visualization.}
  \label{fig:FineCap_450M_1}
\end{figure}

\begin{figure}[h!]
  \centering
  \includegraphics[width=0.95\columnwidth]{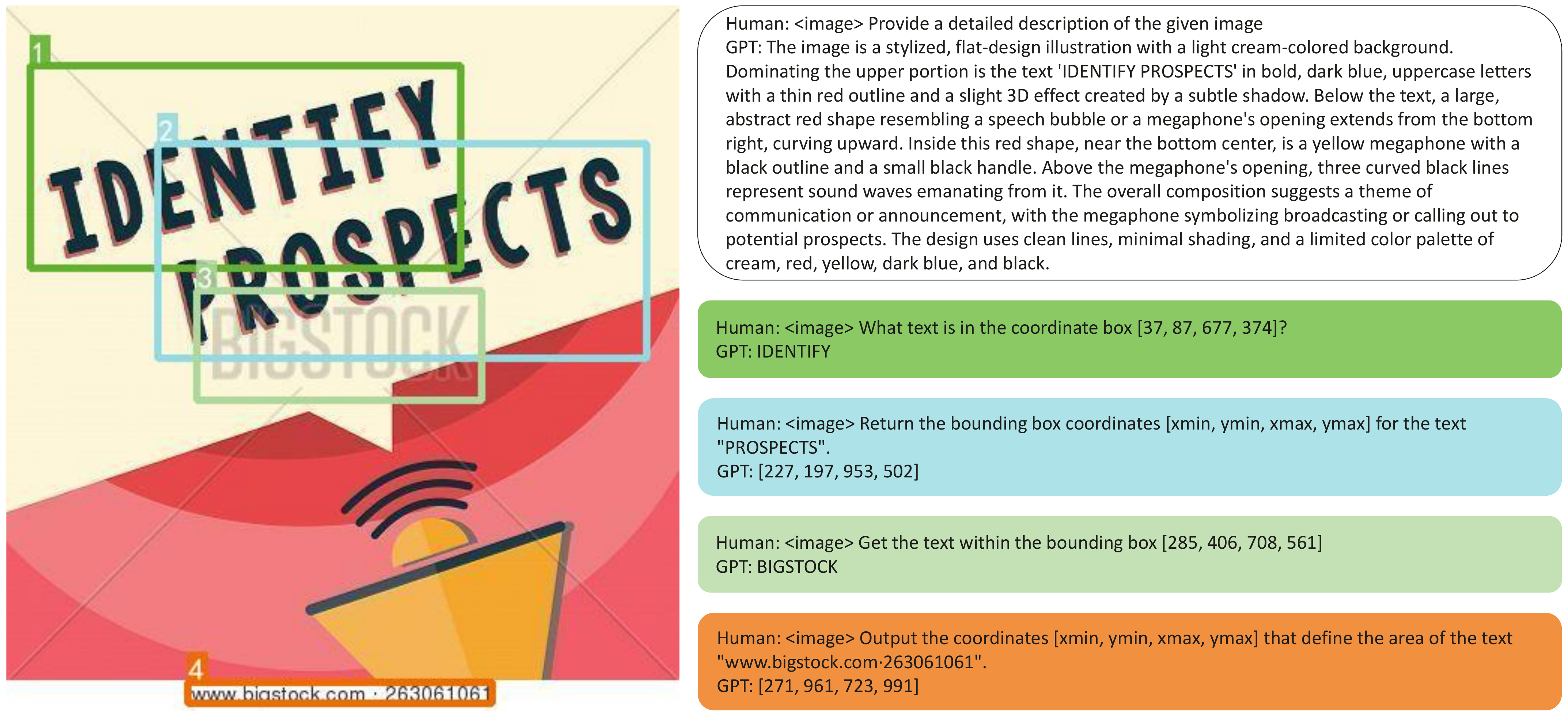}
  \includegraphics[width=0.95\columnwidth]{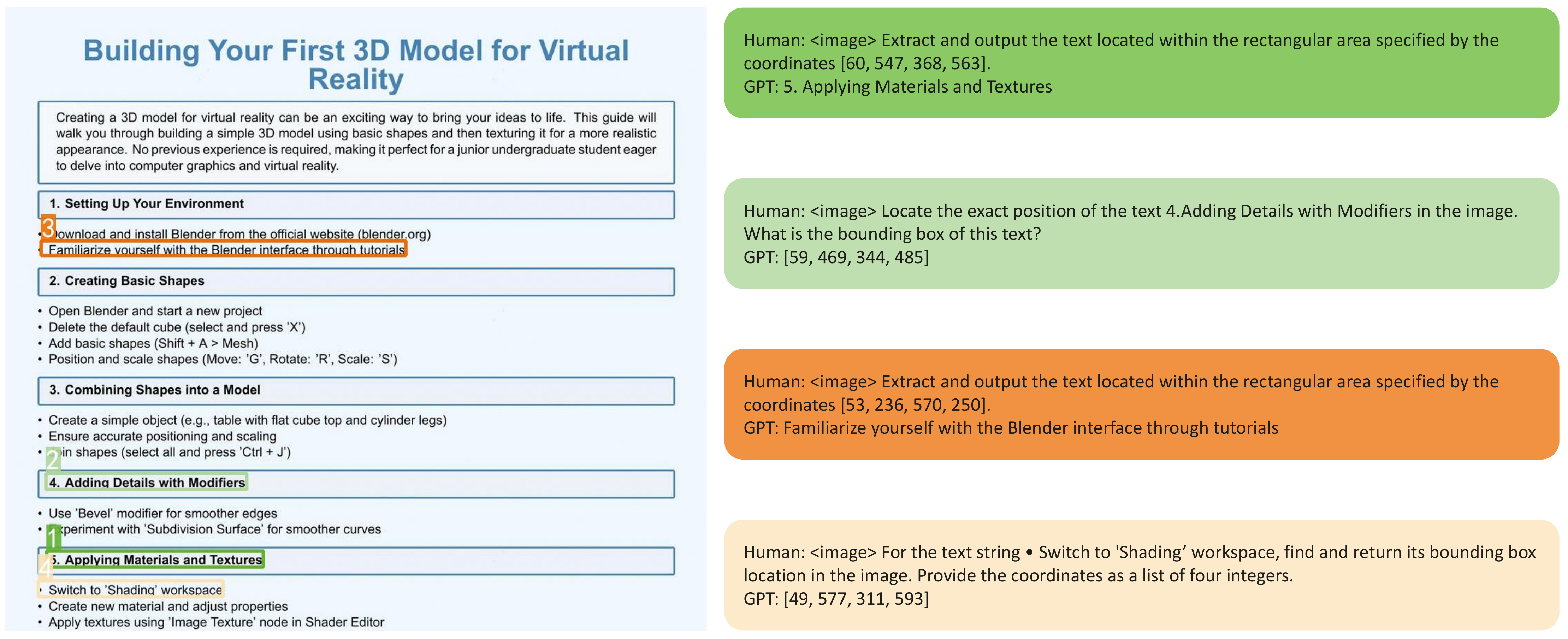}
  \caption{Illustration of FineCap-450M. To prevent an excessive number of regions, we randomly sampled a subset for visualization}
  \label{fig:FineCap_450M_2}
\end{figure}

Figures~\ref{fig:FineCap_450M_1} and~\ref{fig:FineCap_450M_2} present additional samples from the FineCap-450M dataset. In these visualizations, global captions are displayed in white text boxes, while local captions corresponding to specific regions are shown in colored boxes.As illustrated in Figure~\ref{fig:FineCap_450M_1}, which features natural scenes, each key entity is precisely localized using bounding boxes $[x_{min}, y_{min}, x_{max}, y_{max}]$. These annotations span a spectrum from coarse-grained objects (e.g., animals) to fine-grained elements, such as local identification tags, slender mechanical rods, or specific brand logos. Beyond simple category labels, the annotations provide rich descriptions of textures (e.g., "ribbed," "mottled"), materials (e.g., "bamboo," "stainless steel"), and specific physical postures. Consequently, entity localization is driven by relative positioning and descriptive context rather than mere object recognition.For text-dense and document-only images, as shown in Figure~\ref{fig:FineCap_450M_2}, the dataset provides exact coordinate mappings across diverse font styles and hierarchical structures, enabling models to parse and recognize specific textual content with high granularity.

\section{Training for MLLM}

\subsection{Main Results}

\begin{figure}[t]
  \centering
 \begin{minipage}[Ratio of different type of data]{0.49\linewidth}
  \centering
\includegraphics[width=0.95\columnwidth]{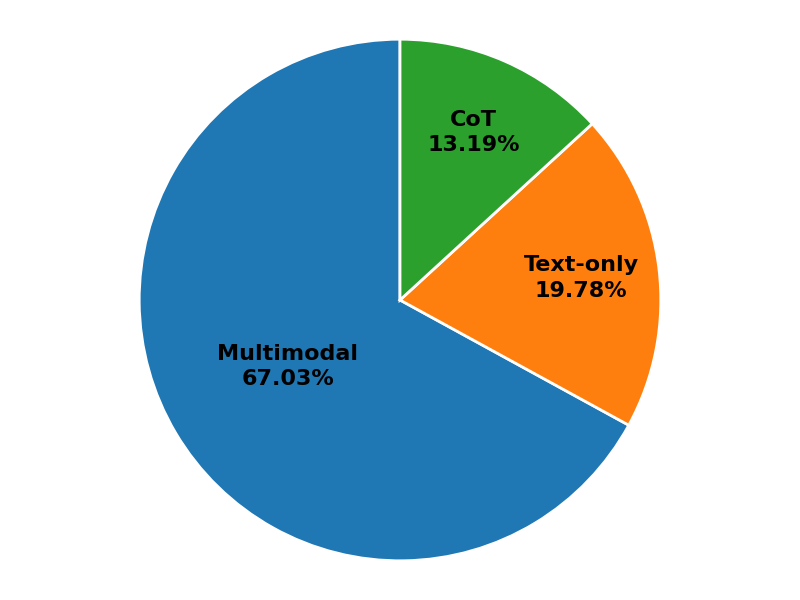}
 \end{minipage}
 \begin{minipage}[Multimodal data distribution]{0.49\linewidth}
  \centering
 \includegraphics[width=0.95\columnwidth]{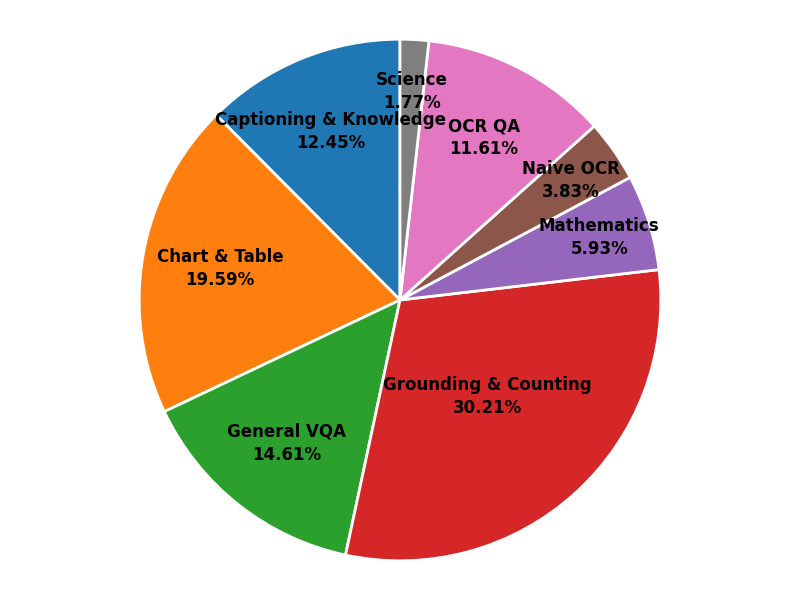}
 \end{minipage}
  \caption{(a) Ratio of different type of data. (b)Multimodal data distribution.}
  \label{fig:data_dist}
\end{figure}

\noindent\textbf{80M SFT Data for MLLM Training.} To endow FineViT with multimodal understanding ability, we curate a large scale of high-quality supervised fine-tuning (SFT) dataset and fine-tune the model after LLM alignment stage. This dedicated SFT dataset consists of approximately 80M QA samples collected from open-source datasets including FineVision~\cite{wiedmann2025finevisionopendataneed}, LLaVA-OneVision-1.5-Instruct~\cite{LLaVA-OneVision-1.5}, Eagle 2~\cite{li2025eagle2buildingposttraining}, Infinity-MM~\cite{gu2025infinitymmscalingmultimodalperformance}, LLaVA-OneVision~\cite{li2024llavaonevisioneasyvisualtask}, and Honey-Data~\cite{zhang2025beehighqualitycorpusfullstack}.  The ratio of different type of data (\textit{i.e.}, text-only, multimodal, and CoT) is shown in Figure~\ref{fig:data_dist}(a). Following FineVision, we also categorize the multimodal QA samples into eight subcategories \textit{i.e.}, Captioning \& Knowledge, Chart \& Table, General VQA, Grounding \& Counting, Mathematics, Naive OCR, OCR QA, Science, and the distribution of different type of multimodal data is illustrated in Figure~\ref{fig:data_dist}(b).

\noindent\textbf{Training Recipe.} To enable a rigorous comparison with other small-scale VLMs, we fine-tuned our model, FineViT-VL, using the curated $80$M SFT dataset following the stage III, LLM alignment stage. The supervised fine-tuning (SFT) is conducted in two consecutive phases. In Phase I, the model is trained with a batch size of $1024$ and a learning rate of $1 \times 10^{-5}$, with the maximum token length capped at $4096$. In Phase II, to enhance the model's capacity for processing extended contexts, we increase the maximum token length to $10240$, while maintaining the learning rate at $1 \times 10^{-5}$ and adjusting the batch size to $512$.


\subsection{Ablation Studies}
In this part, we elaborate the detailed experimental setting for ablation studies presented in the main text.

\noindent\textbf{Comparison of Vision Encoders across Various LLM Scales (Table~\ref{tab:ablation_vit_llm}.)} To make a fair and comprehensive comparison with SigLIP2-naflex (\textit{i.e.}, the native-resolution version of SigLIP2) and better evaluate the effectiveness of vision encoder in MLLM, we freeze the vision encoder during MLLM training and set the maximum resolution of images fed into different ViTs the same \textit{i.e.}, within $1000 \times 1000$. Specifically, 10M InfinityMM~\cite{gu2025infinitymmscalingmultimodalperformance} caption data is firstly used to train a two-layer projector with batch size $512$ and learning rate $1.0 \times 10^{-3}$. Then, 6M SFT randomly sampled from the aforementioned 80M SFT data is adopted to jointly train the projector and LLM, with batch size $256$ and learning rate $1.0 \times 10^{-5}$. We compare the performance vision encoders with various LLM scales \textit{i.e.}, Qwen3-1.7B and Qwen3-8B. 

\noindent\textbf{Effectiveness of Progressive Training Stages (Table~\ref{tab:comparison_different_stages}.)}
We systematically evaluate the contribution of each training phase. To fully unlock the potential of the vision encoder in training MLLM, a three-stage training paradigm~\cite{LLaVA-OneVision-1.5} is adopted. Firstly,  the same 10M InfinityMM caption data is used to train a two-layer MLP projector with batch size $512$ and learning rate $1.0 \times 10^{-3}$. Then, 8M global caption data randomly sampled from the LLaVA-OneVision-1.5-Mid-Traning dataset~\cite{LLaVA-OneVision-1.5} is adopted to jointly train the ViT and projector with batch size $512$ and learning rate $2.0 \times 10^{-5}$. Finally, the same 6M SFT data mentioned above is used for full parameter training with batch size $256$ and learning rate $1.0 \times 10^{-5}$. 

\begin{figure}[h!]
  \centering
  \includegraphics[width=1.0\columnwidth]{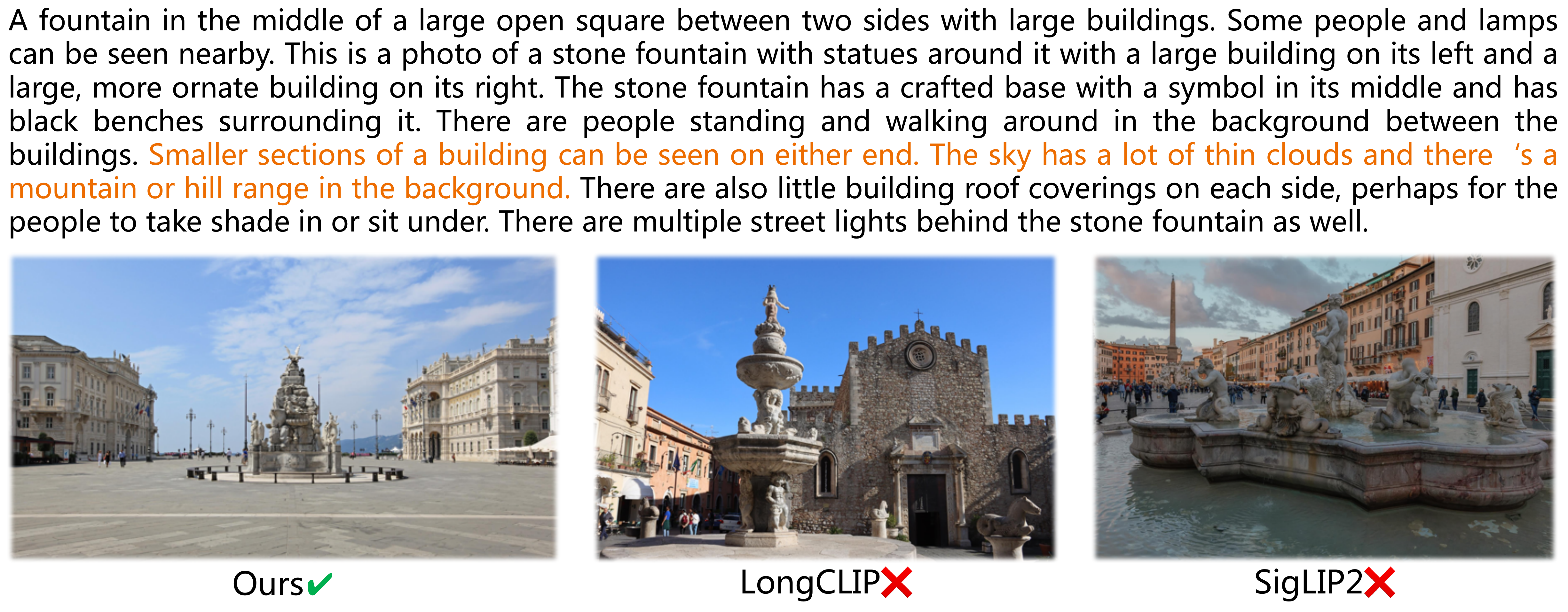}
  \includegraphics[width=1.0\columnwidth]{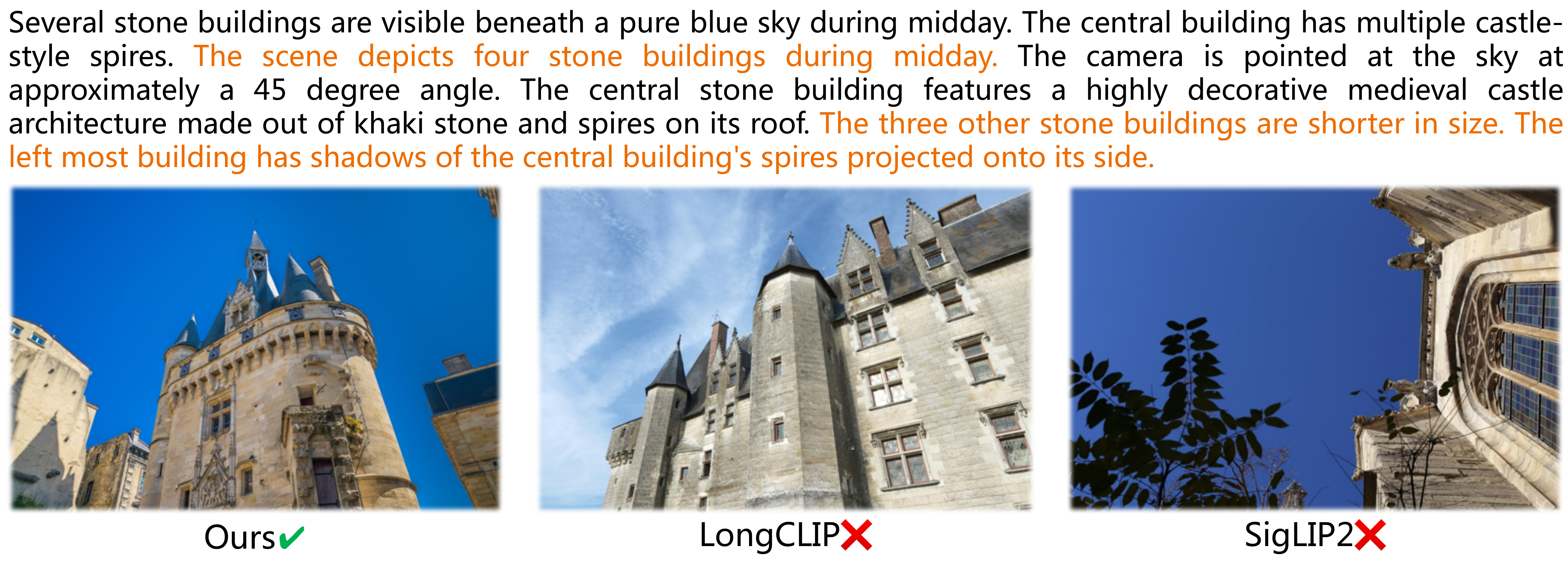}
  \includegraphics[width=1.0\columnwidth]{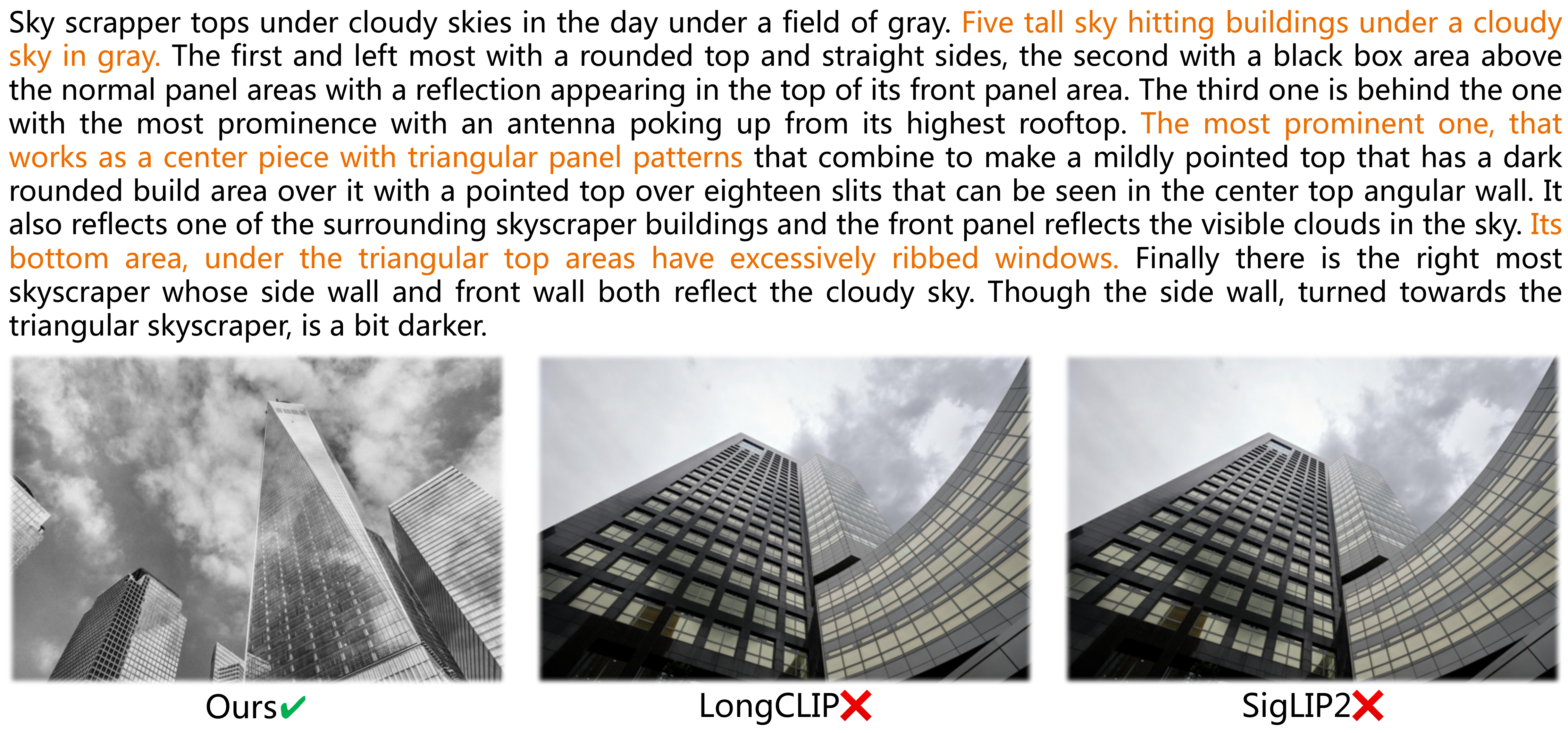}
  \caption{Long-text Zero-shot Retrieval. Distinctive attributes within the long caption are highlighted in {\color{orange}{orange}} to help identify the correct image.}
  \label{fig:long_text_retrieval1}
\end{figure}

\begin{figure}[h!]
  \centering
  \includegraphics[width=1.0\columnwidth]{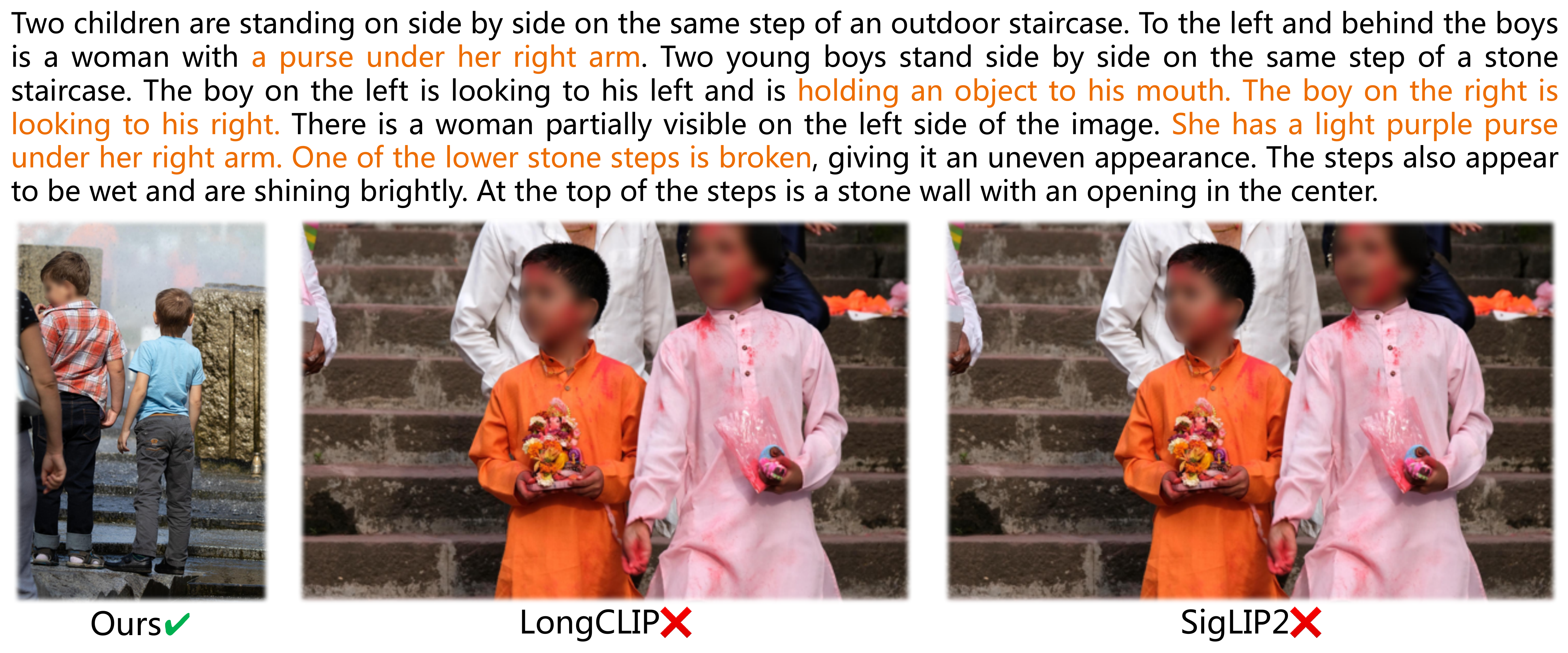}
  \includegraphics[width=1.0\columnwidth]{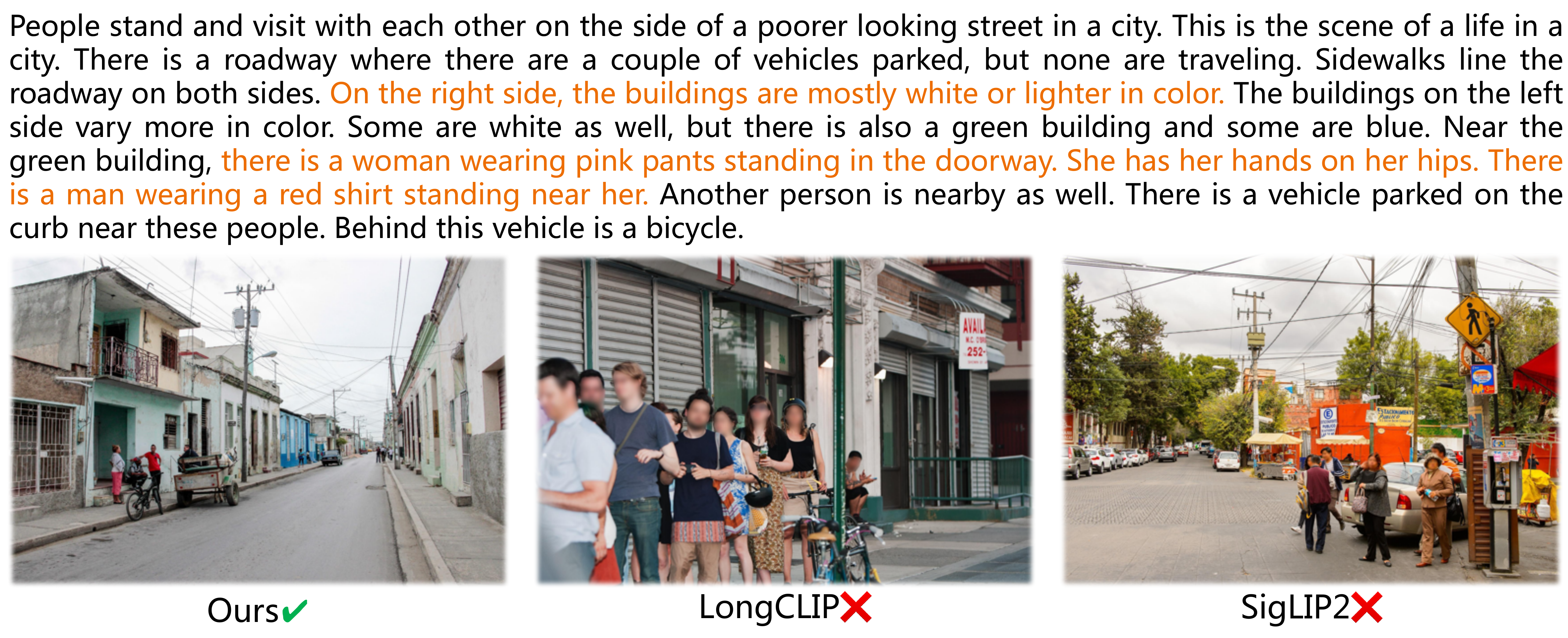}
  \includegraphics[width=1.0\columnwidth]{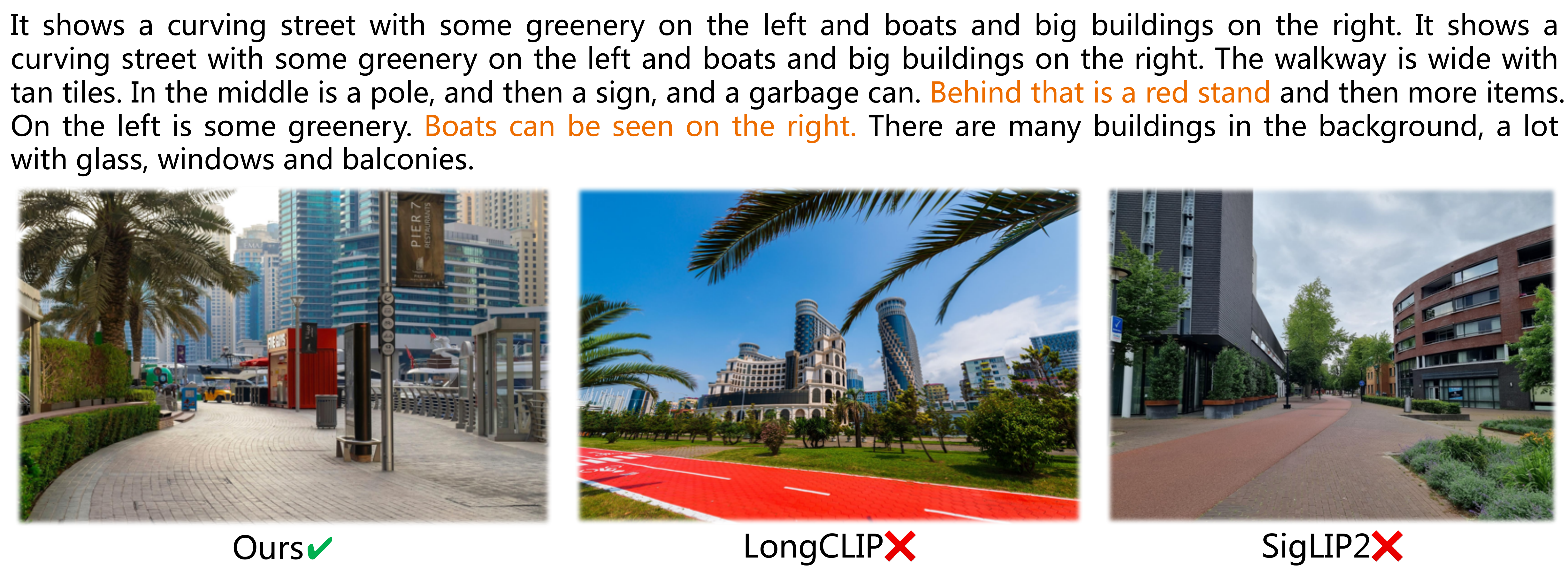}
  \caption{Long-text Zero-shot Retrieval. Distinctive attributes within the long caption are highlighted in {\color{orange}{orange}} to help identify the correct image.}
  \label{fig:long_text_retrieval2}
\end{figure}

\section{More Illustration}
\subsection{Long-text Zero-shot Retrieval}
To further evaluate the performance of FineViT in long-text zero-shot retrieval, we provide additional visual comparisons with other state-of-the-art models (LongCLIP and SigLIP2) in Figure~\ref{fig:long_text_retrieval1} and Figure~\ref{fig:long_text_retrieval2}. As illustrated in the qualitative results, FineViT consistently retrieves the exact matching images by accurately capturing the fine-grained visual details and dense spatial relationships described in multi-sentence prompts. While baseline models often struggle to distinguish between visually similar scenes, FineViT excels at comprehending complex architectural structures (such as medieval spires or reflective skyscrapers) , precise human-object configurations (like two boys standing on a specific stone step) , and nuanced contextual cues within cluttered urban environments (such as specific clothing colors or small street signs). These comparisons underscore our model's superior capability in bridging the gap between lengthy, highly descriptive language and intricate visual features.






\end{document}